%% file: main.tex
\documentclass[11pt]{scaleai-paper}

\usepackage{amsmath}
\usepackage{amsfonts}
\usepackage{amssymb}
\usepackage{booktabs}
\usepackage[square,numbers,sort&compress]{natbib}
\usepackage{nicefrac}
\usepackage{fvextra}
\usepackage{placeins}
\usepackage{tikz}
\usepackage{url}
\usepackage{fontawesome5}
\usepackage[colorlinks=true,linkcolor=scaleLink,citecolor=scaleLink,urlcolor=scaleLink]{hyperref}
\usetikzlibrary{arrows.meta,positioning,fit,backgrounds,calc}

\graphicspath{{figures/}}

\newcommand{\preping}{\textsc{PREPING}}
\newcommand{\corpusToSkill}{\textsc{Corpus2Skill}}
\newcommand{\metaAgent}{\textsc{Meta-Agent}}
\newcommand{\metaNoArchive}{\textsc{Meta-Agent w/o Archive}}
\newcommand{\metaWithArchive}{\textsc{Meta-Agent w/ Archive}}
\newcommand{\noStudy}{\textsc{No Study}}
\definecolor{tablestripe}{HTML}{F5F7FA}
\newsavebox{\bestscorebox}
\newcommand{\bestscore}[1]{%
  \sbox{\bestscorebox}{$#1$}%
  \resizebox{\wd\bestscorebox}{\height}{$\boldsymbol{#1}$}%
}
\newcommand{\secondscore}[1]{\ensuremath{\underline{#1}}}

\title{Studying Without a Syllabus: \\Task-Agnostic Environment Preprocessing}

\author[1,2,*]{Vinay Samuel}
\author[1]{Varun Ursekar}
\author[1]{Vijay S. Kalmath}
\author[1]{Apaar Shanker}
\author[1]{Veronica Chatrath}
\author[1]{Yuan Xue}
\affil[1]{Scale AI}
\affil[2]{University of Maryland, College Park}
\contact{\textit{*Work done during internship at Scale AI}\\[0.25em]
\mbox{\textcolor{scaleBlack}{\faEnvelope}\hspace{0.35em}%
\href{mailto:varun.ursekar@scale.com}{\textcolor{scaleBlack}{\texttt{varun.ursekar@scale.com}}}}%
\hspace{1.25em}%
\mbox{\textcolor{scaleBlack}{\faGithub}\hspace{0.35em}%
\href{https://github.com/scaleapi/meta-study}{\texttt{github.com/scaleapi/meta-study}}}}

\begin{document}

\maketitle

\begin{abstract}
Before an LLM agent tackles tasks in a new environment, it can inspect available corpora and tools and construct reusable resources such as indices, scripts, or procedural guidance. Most automated adaptation methods, however, rely on task examples, trajectories, or evaluation feedback to decide what to build. Existing task-agnostic approaches avoid this supervision but commit in advance to a preparation strategy for a particular type of environment. We study a more open-ended setting: can an agent study an unfamiliar environment without a syllabus, i.e. before test time and without knowledge of the downstream task distribution, and choose how to prepare it? We formalize task-agnostic environment preprocessing, in which a studying system explores an environment under a budget and produces artifacts for a frozen solver. We compare unaided and archive-equipped \metaAgent{}s with fixed synthetic-practice and corpus-processing methods across six heterogeneous benchmarks. A \metaAgent{} variant achieves the highest Avg@3 reward on five benchmarks, while fixed corpus processing remains best on the largest corpus benchmark. Larger study budgets do not reliably improve downstream reward. Nevertheless, studied artifacts reduce the test-time sampling needed to reach a given score, demonstrating how reusable preparation can shift computation from repeated test-time attempts to a pre-task study phase.

\end{abstract}

\input{sections/introduction.tex}

\input{sections/related_work.tex}

\input{sections/formulation.tex}

\input{sections/method.tex}
\input{sections/setup.tex}
\input{sections/results.tex}
\input{sections/qualitative.tex}
\input{sections/discussion.tex}
\input{sections/conclusion.tex}

\newpage
\small
\bibliographystyle{plainnat}
\bibliography{references}


\normalsize
\appendix

\clearpage
\section{Supporting Materials}
This section provides supporting material for our experiment setup, main results, and analysis.

\input{sections/appendix_benchmarks.tex}
\input{sections/appendix_results.tex}
\input{sections/appendix_methods.tex}

\input{sections/appendix_interpretability.tex}

\end{document}

%% file: sections/introduction.tex
\section{Introduction}
\label{sec:intro}

LLM agent performance depends substantially on choices beyond the underlying model: prompts, tools, context and memory management, and the organization of corpora, databases, and external services~\citep{browsergymecosystem,weng2026harness,sweagent, yao2026harnessbenchmeasuringharnesseffects}. Consider an agent that performs retrieval over a large filesystem. If files are unnamed and disorganized, the agent receives no clues about file content and must resort to expensive, exhaustive search. Consequently, practitioners routinely adapt components of the agent \emph{harness} and \emph{environment} to tailor them to the tasks the agent is expected to perform. This has led to increased interest in automated approaches to optimizing agents for novel tasks, spanning prompt optimization, harness optimization, and adaptive memory.

Existing automated adaptation methods typically use information about the expected test-time task distribution to guide modifications. Prompt optimizers such as DSPy~\citep{dspy} and GEPA~\citep{gepa} use \emph{task examples} and \emph{evaluation feedback}. VeRO~\citep{ursekar2026vero} and Meta-Harness~\citep{lee2026metaharness} modify the entire harness as code using similar signals. Adaptive memory systems such as AWM~\citep{awm}, ACE~\citep{ace}, and Dynamic Cheatsheet~\citep{dynamiccheatsheet} extract reusable knowledge and procedures from \emph{task trajectories} with or without \emph{labels}. Such supervisory signals may be unavailable when an agent first encounters a new environment, particularly when representative tasks are costly to obtain. This cold start motivates \emph{task-agnostic adaptation} of agents and environments prior to test time.

Several lines of work have addressed the problem of agent and environment adaptation prior to deployment. One line substitutes knowledge of the true task distribution with self-generated exploratory practice. SPICE~\citep{liu2025spiceselfplaycorpusenvironments} uses corpus-grounded self-play to generate a curriculum for parameter adaptation, while \preping{}~\citep{preping} generates synthetic tasks in tool environments and distills their trajectories into a procedural playbook. In corpus environments, retrieval structures can be prepared offline to make document collections more amenable to search by agents: RAPTOR~\citep{raptor} recursively clusters and summarizes documents, GraphRAG~\citep{graphrag} constructs a knowledge graph and community summaries, and \corpusToSkill{}~\citep{corpus2skill} compiles a corpus into a navigable skill hierarchy. Closest to our framing, Machine Studying~\citep{machinestudying} treats studying as an explicit pre-task process and evaluates how well an agent can construct reusable context from a corpus without downstream tasks.
Each of these methods commits to a strategy for processing the environment depending on its structure. With many methods available and suited to different environment types, we ask whether an \emph{open-ended studying system} can choose how to process an environment as a function of what it contains.

\begin{figure}[t]
\centering
\includegraphics[width=\textwidth]{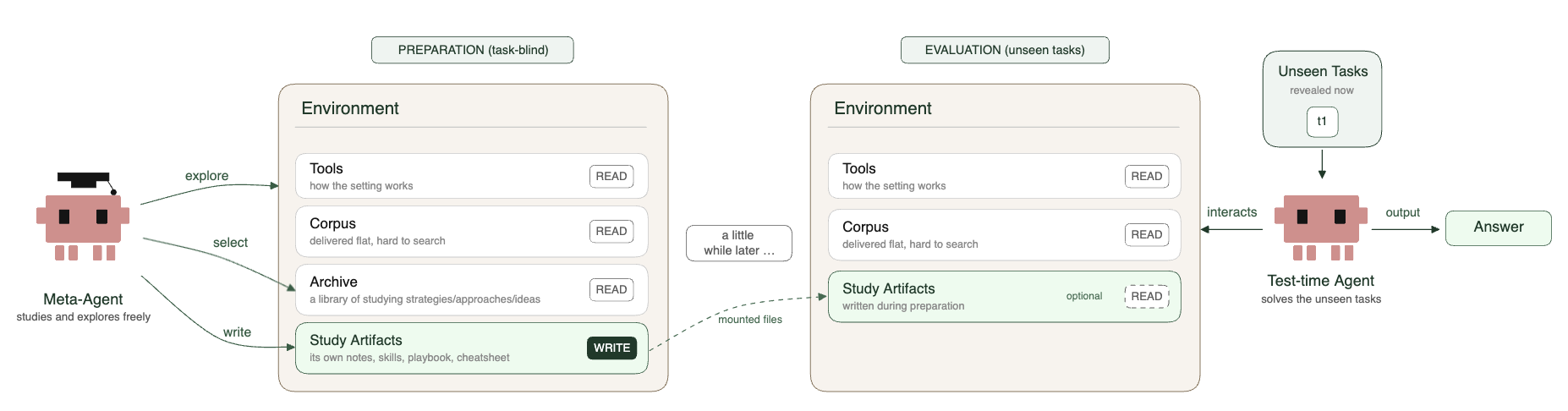}
\caption{An instance of our \metaWithArchive{} pipeline. The environment $E$ is a sandboxed container with a filesystem. The base environment containing benchmark-specific artifacts such as tools and corpora is shared across study and test time. 
Before test time, the meta-agent explores $E$ under budget $B_{\mathrm{study}}$, using strategies from a mounted archive. It writes study artifacts into a volume that is later mounted into the test-time agent's sandbox. The test-time agent uses the study artifacts and the pre-existing benchmark-specific artifacts to solve unseen tasks.}
\label{fig:pipeline}
\end{figure}

Specifically, we ask whether a \metaAgent{} can explore an environment without knowledge of the downstream task distribution and nonetheless produce artifacts that enable a frozen solver agent to effectively perform tasks at test time. The solver's weights and harness are fixed. What changes is the set of resources it finds in its sandboxed environment. These artifacts may extend the environment with directories, indices, knowledge bases, scripts, or tools, or supply harness-side context through prompted guidance, skills, or playbooks. Any file-based artifact the solver can access from within its sandbox is admissible. We compare two variants of this \metaAgent{} against \preping{} and \corpusToSkill{}, two fixed workflows that process agent environments in a task-agnostic way.

Across six benchmarks spanning 36 unique environments, each with differently sized corpora, diverse tools, and specialized domain knowledge, the two \metaAgent{} variants collectively lead to the highest downstream Avg@3 reward on five, underperforming \corpusToSkill{} only on BCP-G. Our archive-equipped variant improves over \noStudy{} on all six, and ranks first or second under both Avg@3 and Best@3 of downstream held-out rewards. We also analyze what the methods explore and produce, how solver performance scales with study budget, and how test-time sampling scales with and without studying.

We make four contributions:
\begin{enumerate}
\item We formalize \textbf{task-agnostic environment processing}: systems receive an environment and a bounded study budget, but no downstream task instances, traces, or labels, and construct reusable environment resources, harness-delivered context, or both for a frozen solver.
\item We introduce \textbf{open-ended studying \metaAgent{}s} that dynamically choose \emph{how to explore} an environment and \emph{what to create}, with unaided and archive-assisted variants.
\item We show that \textbf{open-ended studying strategies outperform fixed ones on five of the six heterogeneous benchmarks} we evaluate on.
\item We show that additional study budget does not reliably improve downstream performance, while studied agents often require less test-time compute to reach a given score than agents without studying.

\end{enumerate}

%% file: sections/related_work.tex
\section{Related Work}
\label{sec:related}

\paragraph{Task-informed adaptation of agent harnesses.}
Downstream task information can guide both what an agent system should change and whether the change helped. DSPy~\citep{dspy}, GEPA~\citep{gepa}, and PromptBreeder~\citep{promptbreeder} optimize instructions or demonstrations against task examples and evaluation signals, while VeRO~\citep{ursekar2026vero} and Meta-Harness~\citep{lee2026metaharness} extend this search to executable harness components. Memory and context systems, including AWM~\citep{awm}, Memp~\citep{memp}, ACE~\citep{ace}, Dynamic Cheatsheet~\citep{dynamiccheatsheet}, Reflexion~\citep{reflexion}, ExpeL~\citep{expel}, MemGPT~\citep{memgpt}, A-MEM~\citep{amem}, and Mem0~\citep{mem0}, instead distill knowledge, procedures, or code from observed interactions. Other work relocates this processing within the interaction lifecycle: \textsc{ProAct}~\citep{proact} acts between user turns, \textsc{IdleSpec}~\citep{idlespec} during tool-call latency, and \textsc{Auto-Dreamer}~\citep{autodreamer} after experience has accumulated. These methods use task examples, trajectories, or feedback to decide what to change. Our setting asks what to construct before such task information is available.

\paragraph{Task-agnostic adaptation of agent harnesses and environments.}
Without task information, a system must derive its objective from what is available before use. Sleep-time Compute~\citep{sleeptime} studies how preprocessing a supplied context can displace computation after an unknown query arrives. Machine Studying~\citep{machinestudying} isolates the same temporal regime as ours (study before future queries) but fixes both the input modality to a corpus and the candidate interventions to self-supervised training, synthetic-data fine-tuning, and amortized cheatsheet construction. Our decision space instead includes heterogeneous environments: the studying system must determine what to process and what form of reusable support to produce. Existing approaches encode different commitments about what will transfer. Retrieval systems assume that useful preparation takes the form of queryable corpus structure, from conventional sparse retrieval such as BM25~\citep{bm25}, dense retrieval such as DPR~\citep{dpr}, and retrieval-augmented generation ~\citep{rag} to the summaries, graphs, and hierarchies constructed by RAPTOR~\citep{raptor}, GraphRAG~\citep{graphrag}, \corpusToSkill{}~\citep{corpus2skill}, and \textsc{PANINI}~\citep{panini}. Practice-based systems instead rely on self-generated interaction: Voyager~\citep{voyager} acquires executable skills through open-ended exploration, SPICE~\citep{liu2025spiceselfplaycorpusenvironments} constructs a corpus-grounded training curriculum, and \preping{}~\citep{preping} practices with tools to produce a procedural playbook. \preping{}'s guided-exploration baseline is the nearest operational analogue to \metaNoArchive{}: both explore before target tasks and record reusable guidance. We place these commitments under a common protocol and test whether a study agent can choose and combine forms of preparation based on the environment.

%% file: sections/formulation.tex
\begin{table}[t]
\centering
\small
\begingroup
\setlength{\tabcolsep}{5.5pt}
\renewcommand{\arraystretch}{1.16}
\resizebox{\textwidth}{!}{%
\begin{tabular}{@{}lllllll@{}}
\toprule
\textbf{Method} & \multicolumn{1}{c}{\textbf{BCP-G}} & \multicolumn{1}{c}{\textbf{OfficeQA}} & \multicolumn{1}{c}{\textbf{Harvey LAB}} & \multicolumn{1}{c}{\textbf{DABStep}} & \multicolumn{1}{c}{\textbf{Apex Agents}} & \multicolumn{1}{c}{\textbf{AppWorld}} \\
\midrule
\noStudy{}               & $0.197 \pm 0.007$ & $0.549 \pm 0.017$ & $0.260 \pm 0.007$ & $0.353 \pm 0.008$ & $0.183 \pm 0.004$ & $0.576 \pm 0.013$ \\
\addlinespace[2pt]
\rowcolor{tablestripe}
\preping{}               & $0.285 \pm 0.012$ & $0.609 \pm 0.020$ & $0.299 \pm 0.013$ & $\secondscore{0.464 \pm 0.060}$ & $\secondscore{0.231 \pm 0.010}$ & $0.645 \pm 0.036$ \\
\corpusToSkill{}         & \bestscore{0.471 \pm 0.015} & $0.535 \pm 0.029$ & $0.227 \pm 0.006$ & $0.324 \pm 0.009$ & $0.221 \pm 0.001$ & $0.603 \pm 0.028$ \\
\addlinespace[2pt]
\rowcolor{tablestripe}
\metaNoArchive{}         & $0.247 \pm 0.016$ & \bestscore{0.633 \pm 0.014} & $\secondscore{0.361 \pm 0.007}$ & $0.400 \pm 0.006$ & $0.224 \pm 0.004$ & \bestscore{0.688 \pm 0.057} \\
\metaWithArchive{}       & $\secondscore{0.286 \pm 0.021}$ & $\secondscore{0.615 \pm 0.004}$ & \bestscore{0.366 \pm 0.012} & \bestscore{0.529 \pm 0.097} & \bestscore{0.233 \pm 0.011} & $\secondscore{0.665 \pm 0.042}$ \\
\bottomrule
\end{tabular}}
\endgroup
\caption{\textbf{Avg@3 downstream reward.} Best values are bold. Second best are underlined. For each studying method, each artifact-iteration score averages three task-time repetitions per task. Values report the mean $\pm$ sample SD across three independent artifact iterations. For \noStudy{}, values report the mean $\pm$ population SD across all $\binom{10}{3}$ three-repetition subset estimates.}
\label{tab:main-avg}
\end{table}  

\begin{table}[t]
\centering
\small
\begingroup
\setlength{\tabcolsep}{5.5pt}
\renewcommand{\arraystretch}{1.16}
\resizebox{\textwidth}{!}{%
\begin{tabular}{@{}lllllll@{}}
\toprule
\textbf{Method} & \multicolumn{1}{c}{\textbf{BCP-G}} & \multicolumn{1}{c}{\textbf{OfficeQA}} & \multicolumn{1}{c}{\textbf{Harvey LAB}} & \multicolumn{1}{c}{\textbf{DABStep}} & \multicolumn{1}{c}{\textbf{Apex Agents}} & \multicolumn{1}{c}{\textbf{AppWorld}} \\
\midrule
\noStudy{}               & $0.350 \pm 0.014$ & $0.730 \pm 0.017$ & $0.398 \pm 0.011$ & $0.487 \pm 0.015$ & $0.297 \pm 0.009$ & $0.685 \pm 0.013$ \\
\addlinespace[2pt]
\rowcolor{tablestripe}
\preping{}               & $0.443 \pm 0.013$ & $0.747 \pm 0.016$ & $0.426 \pm 0.017$ & $\secondscore{0.565 \pm 0.042}$ & $\secondscore{0.342 \pm 0.010}$ & $0.760 \pm 0.040$ \\
\corpusToSkill{}         & \bestscore{0.680 \pm 0.017} & $0.717 \pm 0.010$ & $0.322 \pm 0.010$ & $0.446 \pm 0.005$ & $0.331 \pm 0.005$ & $0.710 \pm 0.018$ \\
\addlinespace[2pt]
\rowcolor{tablestripe}
\metaNoArchive{}         & $0.410 \pm 0.019$ & \bestscore{0.780 \pm 0.016} & $\secondscore{0.485 \pm 0.009}$ & $0.549 \pm 0.012$ & $0.329 \pm 0.004$ & \bestscore{0.780 \pm 0.059} \\
\metaWithArchive{}       & $\secondscore{0.451 \pm 0.023}$ & $\secondscore{0.749 \pm 0.006}$ & \bestscore{0.488 \pm 0.008} & \bestscore{0.653 \pm 0.088} & \bestscore{0.343 \pm 0.017} & $\secondscore{0.778 \pm 0.033}$ \\
\bottomrule
\end{tabular}}
\endgroup
\caption{\textbf{Best@3 downstream reward.} Best values are bold and second-best values are underlined. For each studying method, each artifact-iteration score takes the best of three task-time repetitions per task. Values report the mean $\pm$ sample SD across three independent artifact iterations. For \noStudy{}, values report the mean $\pm$ population SD across all $\binom{10}{3}$ three-repetition subset estimates. Best@3 consumes three task-time rollouts and is not a fixed-cost substitute for Avg@3.}
\label{tab:main-best}
\end{table}

\section{Task-Agnostic Environment Preprocessing}
\label{sec:formulation}
\label{sec:axes}
\label{sec:task}

In our setup, an agent $\pi$ operates in an environment $E$. An agent $\pi = (m, H)$ is a tuple of an LLM $m$ and a harness $H$. The environment $E$ is a shared runtime and collection of resources required to complete a set of tasks. Practically, $E$ is implemented as an isolated Docker sandbox compatible with the Harbor framework~\citep{Harbor_Framework}.  The harness $H$ represents the program that invokes the model within the environment. $H$ maintains context, registers and invokes tools, and drives the interaction loop between the model and environment. While any standard CLI-based coding harness is compatible with our setup, we use the Claude Code~\citep{anthropic_claude_code_docs} harness throughout this work.

Each environment $E$ admits a space $\mathcal{T}_E$ of feasible downstream tasks. Each task $t = (x_t, r_t) \in \mathcal{T}_E$ consists of a prompt $x_t$ and a reward function $r_t$ that scores the output of an agent prompted with $x_t$. A benchmark provides a finite held-out test set; we write $D_E$ for the uniform empirical distribution over that set. Let $\pi_{\mathrm{solver}}$ denote a frozen downstream agent. Its aggregate reward on an environment $E'$ and test distribution $D_E$ is then
\begin{equation}
R_{\pi_{\mathrm{solver}}}(E', D_E) = \mathbb{E}_{t \sim D_E}
\left[ r_t\!\left( \pi_{\mathrm{solver}}(x_t, E') \right) \right],
\label{eq:agg_reward}
\end{equation}
where $E'$ is either the original environment $E$ or a version of it modified before test time.
 
The goal of a task-agnostic \emph{studying system} $S$ is to improve $R_{\pi_{\mathrm{solver}}}(E, D_E)$ without observing $D_E$. In our work, $S$ does this by using the frozen solver and the original environment to output a modified environment $E_{\mathrm{studied}}$, i.e.\ $S : \Pi \times \mathcal{E} \rightarrow \mathcal{E}$, where $\Pi$ and $\mathcal{E}$ are the spaces of agents and environments, respectively. We hold the implementation of the harness $H$ fixed, although artifacts may be supplied through its existing context interfaces. Here, task-agnostic means that $S$ does not receive real downstream task instances or signals derived from their distribution, including traces, labels, verifier outputs, or evaluation feedback. It may interact with $E$ and use records of that interaction, including synthetic practice, to guide its modifications. $S$ may produce any number and type of artifacts materialized as files in $E_{\mathrm{studied}}$, such as indices, knowledge bases, scripts, or executable tools. $E_{\mathrm{studied}}$ contains all of the original artifacts in $E$ together with any additional artifacts intended to help the solver. This preserves the integrity of the original environment.

The studying process incurs a cost $C_{\mathrm{study}}(S, E)$, measured in API dollars, that must not exceed a given study budget $B_{\mathrm{study}}$. Let $p$ denote a distribution over environments. We seek a studying system that maximizes the expected aggregate reward of the frozen solver across environments:
\begin{equation}
S^\star = \arg\max_{S}\;
\mathbb{E}_{E \sim p}\!\left[
R_{\pi_{\mathrm{solver}}}\!\left(S(\pi_{\mathrm{solver}}, E),\, D_E\right)
\right]
\quad \text{s.t.} \quad
C_{\mathrm{study}}(S, E) \leq B_{\mathrm{study}} \;\;.
\label{eq:objective}
\end{equation}
There is no a priori restriction on the structure or logic of $S$. It may be a filesystem processing pipeline as in \corpusToSkill{}, a fixed orchestration of several specialized agents as in \preping{}, or an open-ended agentic process like our \metaAgent{} variants. 




%% file: sections/method.tex
\input{figures/fig_study_scaling.tex}

\section{Methods}
\label{sec:method}

In practice, what $S$ should do depends on the environment.  A large corpus may benefit from indexing and summarization, whereas a broad tool surface may require skills or instructions distilled from synthetic practice or self-play. Figure~\ref{fig:spectrum} describes the breadth of environments in our benchmark suite using two observable measures: counts of files exposed to the solver and benchmark-specific tools. 

One of our aims is to compare the effectiveness of open-ended study strategies versus fixed ones. We call a studying system open-ended when its studying procedure and artifact type are not fixed in advance. We consider two representative baseline systems with contrasting but fixed adaptation procedures, each designed for different environmental niches: \preping{} and \corpusToSkill{}. In contrast, our two \metaAgent{} variants use a single study agent $\pi_{\mathrm{study}}$ that can simultaneously explore an environment and modify it in an open-ended way. We provide a description of all methods below with additional implementation details reported in Appendix~\ref{app:implementation-details}.

\subsection{Baselines: Fixed studying systems}
\paragraph{\preping{}.} 
\preping{} \citep{preping} performs task-agnostic environment preparation through environment-grounded synthetic practice organized into cycles. In each cycle, a \emph{Proposer} generates a batch of tasks based on available tools, a practice \emph{Solver} attempts them in the target harness, and a \emph{Validator} filters infeasible trajectories. A reflector and curator then distill lessons from successes and failures into a procedural playbook containing strategies, pitfalls, interface notes, and reusable code. The number of cycles, tasks per cycle, and per-task execution limit determine the amount of synthetic practice and the overall study cost. At test time, a configurable number of playbook entries are retrieved and prepended to the task instruction. Thus, although \preping{} uses agents, its roles, invocation order, and output format are fixed.

\paragraph{\corpusToSkill{}.}
\corpusToSkill{} \citep{corpus2skill} restructures a document corpus into a navigable directory of skills. Its pipeline summarizes and embeds documents, clusters them into a labeled hierarchy, and constructs indexes that map topics to source documents. The pipeline is not agentic: it follows a fixed end-to-end procedure without adapting its strategy in response to runtime feedback. The amount of source content processed and the structure of the resulting hierarchy are controlled by a document-length limit, whether document summaries are generated, the hierarchy's branching ratio, the maximum number of top-level clusters, and the minimum cluster size. These choices affect the studying cost in turn. At task time, the downstream agent receives navigation instructions, as well as the generated hierarchical filesystem. Unlike \preping{}, this system compiles declarative content without generating or executing synthetic tasks.

\subsection{Open-ended studying systems}
Both \metaAgent{} variants instantiate a study agent $\pi_{\mathrm{study}}$. Under the same task-agnostic protocol, the study agent must explore the environment, choose a studying procedure itself, and construct any file-based artifacts it deems useful to a future solver. The unaided variant \metaNoArchive{} must choose without guidance. \metaWithArchive{} is additionally given a seed archive of skills comprising executable scripts and their descriptions. The archive includes skills implementing \preping{} and \corpusToSkill{}, plus a general exploratory-study workflow resembling \metaNoArchive{}. The skills are mounted into the agent sandbox's filesystem and are accessible using standard shell-based tools (see Figure~\ref{fig:pipeline}). The agent may invoke these workflows during study and use, combine, or ignore their outputs when constructing the final artifacts.

%% file: figures/fig_study_scaling.tex
\begin{figure}[t]
\centering
\includegraphics[width=\textwidth]{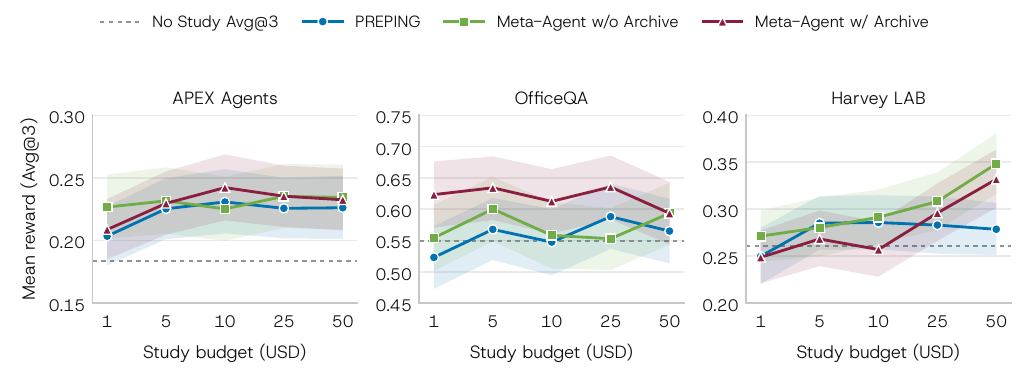}
\caption{\textbf{Study-budget scaling.} Mean Avg@3 reward versus study budget. Shading shows 95\% task-bootstrap intervals, and dotted lines show \noStudy{}. Harvey LAB uses dense verifier pass rate.}
\label{fig:study-scaling}
\end{figure}

%% file: sections/setup.tex
\input{figures/fig_budget.tex}

\section{Experimental Setup}
\label{sec:setup}

\paragraph{Benchmarks.}
We evaluate on six agentic benchmarks spanning diverse environment types. \textbf{BCP-Grep (BCP-G)} adapts BrowseComp-Plus~\citep{browsecompplus}: it retains the 830 questions and fixed 100{,}195-document corpus, but exposes each document as a file for shell-based search rather than through the original BM25 retriever. \textbf{OfficeQA}~\citep{officeqa} and \textbf{Harvey LAB}~\citep{harveylab} also expose document collections. For Harvey LAB, we use only its 250 firm-knowledge tasks, which share the same fictional firm's document-management system. \textbf{DABStep}~\citep{dabstep} exposes seven files that combine task data, answer-bearing references, and protocol instructions for processing that data using Python. \textbf{APEX-Agents}~\citep{apexagents} contributes 452 tasks grouped into 31 worlds, each with its own shared filesystem and tools: eight investment-banking, 12 law, and 11 management-consulting worlds. We treat each world as a separate environment and run the study process independently within each world. Thus, the reported APEX-Agents benchmark score aggregates results across 31 studied environments, whereas each other benchmark represents a single studied environment. \textbf{AppWorld}~\citep{appworld} exposes a multi-application tool interface. Figure~\ref{fig:spectrum} and Table~\ref{tab:bench} report the observed corpora and tools.

\paragraph{Frozen Solver Agent.} For all target task evaluations, we set the frozen solver $\pi_{\text{solver}}$ to an agent that uses the Claude Code harness with Claude Haiku~4.5 as the underlying model at temperature 1.0. 
 
\paragraph{Study Methods.}
Per \S\ref{sec:method}, we compare four study methods \textbf{\preping{}}, \textbf{\corpusToSkill{}}, \textbf{\metaNoArchive{}}, and \textbf{\metaWithArchive{}} against a single \textbf{\noStudy{}} baseline. We use Claude Code as the coding harness for all agentic components. Claude Opus~4.8 performs \metaAgent{} study, as well as \preping{} proposal, validation, reflection, and curation. Claude Haiku~4.5 executes \preping{} synthetic tasks, every downstream task, and \corpusToSkill{} document-card generation. Claude Sonnet~4.6 performs \corpusToSkill{} cluster summarization, labeling, repartitioning, and entity extraction. For both reproduced baselines, we retain the embedding models and embedding-related settings of the original works: \corpusToSkill{} uses \texttt{Qwen3-Embedding-8B} for document and summary embeddings, and \preping{} uses \texttt{text-embedding-3-small} to retrieve playbook entries.

For the primary comparisons, we run \preping{} for five cycles of 10 synthetic tasks, yielding 50 practice tasks per environment; the original implementation uses 10 cycles of 10. We otherwise retain its published validation thresholds. For \corpusToSkill{}, we use the published default configuration where possible, with benchmark-specific adjustments to hierarchy branching, document length, and tree compaction. Additional implementation details and representative prompts appear in Appendices~\ref{app:implementation-details} and~\ref{app:representative-prompts}, respectively.

\paragraph{Repetitions and rollout metrics.}
Each studying system is run independently $N_{\mathrm{study}}=3$ times per environment, producing three artifact sets. For APEX-Agents, this means three study runs for each of its 31 world environments. For every artifact set, downstream evaluation is independently repeated $N_{\mathrm{eval}}=3$ times. We reserve $K$ for the number of those task-time repeats aggregated by Avg@$K$ or Best@$K$. The primary tables use $K=3$. Let $D_E$ be the empirical test-task distribution defined in \S\ref{sec:formulation}, and let $r_{s,t,j}$ be the reward from rollout $j$ on task $t$ using the artifact produced by study iteration $s$. We report
\begin{equation}
\begin{aligned}
 \mathrm{Avg@}K
 &=\frac{1}{N_{\mathrm{study}}K}
   \sum_{s=1}^{N_{\mathrm{study}}}\mathbb{E}_{t\sim D_E}
   \left[\sum_{j=1}^{K}r_{s,t,j}\right], \\
 \mathrm{Best@}K
 &=\frac{1}{N_{\mathrm{study}}}
   \sum_{s=1}^{N_{\mathrm{study}}}\mathbb{E}_{t\sim D_E}
   \left[\max_{1\leq j\leq K}r_{s,t,j}\right]
\end{aligned}
\end{equation}
Avg@$K$ averages all $K$ rollout rewards per task, whereas Best@$K$ selects the best of the $K$ rollouts per task. Both then average over tasks and study iterations. We report the sample standard deviation across the $N_{\mathrm{study}}$ study-run-specific task macro-averages, so uncertainty reflects variation across study runs rather than individual tasks or rollouts.

\noStudy{} has no artifact-set dimension and is evaluated with $N_{\text{eval}}=10$ independent task-time repetitions. For $K=3$, both Avg@3 and Best@3 are computed over all $\binom{10}{3}=120$ subsets of three repetitions: Avg@3 averages the subset means, whereas Best@3 averages the subset maxima. We macro-average over tasks within each subset and report the mean and population standard deviation across the 120 resulting benchmark-level estimates.

\paragraph{Metrics.}
For each target task, we use the benchmark's original evaluation measure except on Harvey LAB. For Harvey LAB, the original task reward is a strict all-criteria-pass indicator over per-task rubrics. We report instead the dense rubric criterion pass rate. The three primary metrics are (i) study cost $C_{\mathrm{study}}$ in API dollars, (ii) downstream reward $r$, and (iii) downstream inference cost $C_{\mathrm{inference}}$ in API dollars. We meter study and downstream inference separately using provider-reported usage and cost where available. Study cost includes all model calls used to construct an artifact, including generation, control, embeddings, and compilation.

\paragraph{Budgets.}
%
For our experiments in Tables~\ref{tab:main-avg} and~\ref{tab:main-best}, we do not impose a shared study budget $B_{\mathrm{study}}$. Instead, each method uses its full configuration, providing a comparison in which study budget is not the limiting constraint. Study costs therefore vary with the method, environment, as well as selected archive workflows in the case of \metaWithArchive{}; Apex Agents costs additionally sum preparation across its 31 independently studied worlds. Separately, we set $B_{\mathrm{study}} \in \{\$1, \$5, \$10, \$25, \$50\}$ for the study-budget scaling experiments.

%% file: figures/fig_budget.tex
\begin{figure}[t]
\centering
\includegraphics[width=\textwidth]{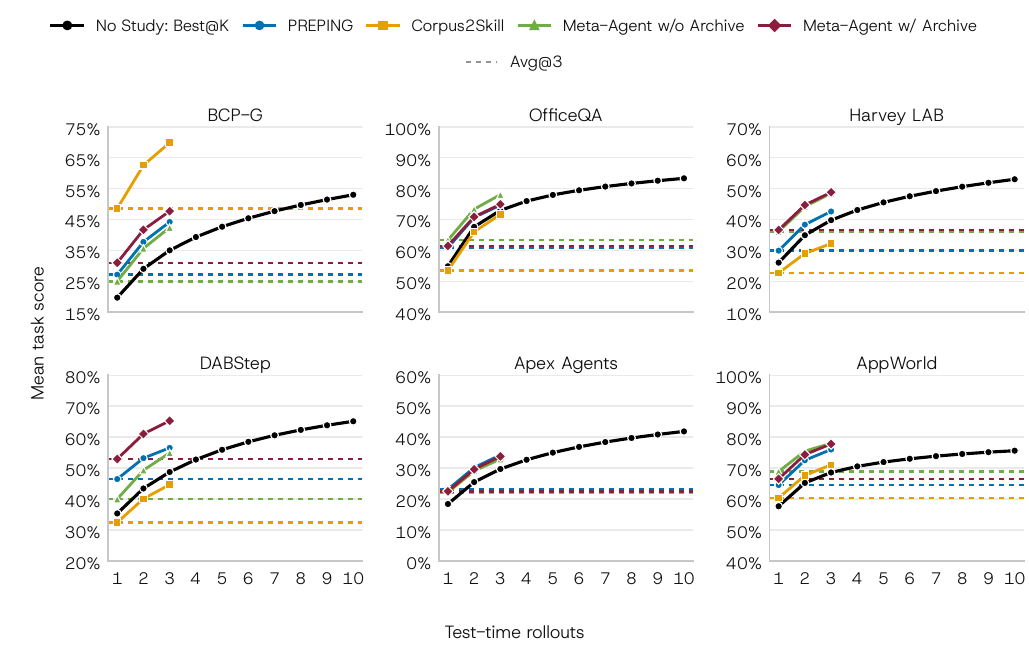}
\caption{\textbf{Test-time scaling with and without study.} \noStudy{} Best@$K$ curves are compared with Best@$K$ curves and Avg@3 reference levels for each study method. Best@$K$ is an oracle upper bound. Harvey LAB uses dense verifier pass rate.}
\label{fig:budget}
\end{figure}

%% file: sections/results.tex
\section{Results}
\label{sec:results}

\paragraph{RQ1: How do studying methods compare across environments?}
Open-ended \metaAgent{} studying is the strongest method family across environments. A \metaAgent{} variant achieves the highest Avg@3 and Best@3 score on 5 of 6 benchmarks, while \metaWithArchive{} ranks first or second on every benchmark under both metrics. Among the committed strategies, \preping{} is consistently beneficial: it exceeds \noStudy{} on all 6 benchmarks under both metrics, although it never ranks first. \corpusToSkill{} exhibits sharper specialization: it is strongest on BCP-G but falls below \noStudy{} on OfficeQA, Harvey LAB, and DABStep. Thus, the \metaAgent{}s combine strong performance with robustness across heterogeneous settings, \preping{} provides smaller but consistent gains, and \corpusToSkill{} trades broader reliability for strong corpus-specific performance. Benchmark-level inference and pre-task costs appear in Tables~\ref{tab:inference-cost-by-benchmark} and~\ref{tab:pretask-cost-by-benchmark}.

\paragraph{RQ2: What is the value of the workflow archive?}
Archive access for the \metaAgent{} has a positive but uneven effect. Under Avg@3, its mean improvement on the four benchmarks where it helps is $0.046$, compared with a mean decline of $0.021$ on the two where it hurts (Table~\ref{tab:main-avg}). The largest gain is $0.129$ on DABStep, whereas the largest decline is $0.023$. Best@3 shows the same asymmetry: a mean gain of $0.041$ where the archive helps and a mean loss of $0.017$ where it hurts (Table~\ref{tab:main-best}). The archive therefore produces occasional large improvements without comparably large regressions, rather than a uniform lift across environments. 

\paragraph{RQ3: How does performance scale with study budget?}
We conduct study budget scaling experiments on OfficeQA, Harvey LAB, and Apex Agents for \preping{}, \metaNoArchive{}, and \metaWithArchive{} for $B_{\mathrm{study}} \in \{\$1, \$5, \$10, \$25, \$50\}$. Results are shown in Figure~\ref{fig:study-scaling}. Additional study budget does not generally improve downstream performance. On OfficeQA and Apex Agents, no method exhibits a sustained increase from \$1 to \$50. The curves are flat or non-monotonic, and most intervals overlap. Harvey LAB is the clear exception. \metaNoArchive{} rises from $0.271$ to $0.348$ and \metaWithArchive{} from $0.249$ to $0.332$; \preping{} remains roughly flat after \$5. Thus, in these experiments, study-budget scaling appears only on Harvey LAB and only for the \metaAgent{} methods.

\paragraph{RQ4: How does test-time compute scale with and without study?}
Studying can reduce the test-time sampling required to reach a given score (Figure~\ref{fig:budget}). The clearest comparison is Best@1 for agents with study artifacts, which uses one rollout per task and requires no oracle selection. To exceed the strongest studied Best@1 score, \noStudy{} requires 8 rollouts on BCP-G, 2 on OfficeQA, 3 on Harvey LAB, 5 on DABStep, 2 on Apex Agents, and 4 on AppWorld. Because \noStudy{} Best@$K$ assumes oracle selection among these outputs, these crossover points are optimistic for \noStudy{} test-time scaling. Using the average per-rollout inference costs in Table~\ref{tab:inference-cost-by-benchmark}, reaching them costs \noStudy{} $1.6$--$5.5\times$ as much task-time inference as the corresponding studied rollout. This comparison excludes the upfront study cost: studying yields a total cost saving only when its artifacts are reused across enough downstream tasks for that cost to amortize. At the higher Best@3 target for studied agents, \noStudy{} requires 6 rollouts on OfficeQA, 7 on Harvey LAB, and 5 on Apex Agents, and does not reach the strongest studied score within 10 rollouts on BCP-G, DABStep, or AppWorld.

%% file: sections/qualitative.tex
\section{Interpretability}
\label{sec:qualitative}

We examine how systems study and when their artifacts help or hinder the downstream solver. Appendices~\ref{app:pretask-interpretability} and~\ref{app:test-interpretability} provide the full protocols, tables, figures, and examples.

\subsection{Studying Process}

A natural hypothesis is that broader exploration helps a studying system distill an environment. We test this by relating downstream reward to counts of distinct files accessed and benchmark-specific tools invoked during study. Appendix~\ref{app:study-coverage} defines the measurement protocol and its limitations.

\textbf{Broader exploration does not reliably predict quality.} Figures~\ref{fig:study-files-vs-score} and~\ref{fig:study-tools-vs-score} show inconsistent relationships with downstream reward. On OfficeQA, \corpusToSkill{} processes all 697 files yet trails methods that inspect fewer. Harvey LAB forms the clearest split: both meta-agents discover its sole benchmark-specific tool, \texttt{labread}, and score 0.35--0.38, while methods that do not invoke it score 0.22--0.31.

\textbf{Exploration strategies vary by environment.} We classify \metaAgent{} study traces to identify recurring behaviors. Appendix~\ref{app:meta-taxonomies} provides the behavior taxonomy and extraction protocol. Table~\ref{tab:study-action-taxonomy} shows that inventory, unavailable-resource tracking, and budget-conscious extraction are common across environments. Test anticipation concentrates in Apex Agents and AppWorld, where histories and examples foreshadow future tasks. Parallel delegation is observed only in Harvey LAB and Apex Agents. The agents combine a shared diagnostic core with environment-specific exploration.

\textbf{Archive use is environment-sensitive but imperfect.} The archive-assisted meta-agent can combine artifacts from open-ended study, \preping{}, and \corpusToSkill{}. The resulting composition varies across environments: open-ended-study and \preping{} outputs appear in every BCP-G, OfficeQA, and DABStep artifact set, whereas \corpusToSkill{} appears in every AppWorld artifact set and 86 of 93 Apex Agents artifact sets (Table~\ref{tab:archive-workflow-usage}). On BCP-G, however, the \metaAgent{} omits \corpusToSkill{} even though it is the strongest standalone method. Archive access therefore supports adaptive composition without guaranteeing the best choice.

\subsection{Solver Process}

\textbf{Benefits concentrate in particular task strata.} On DABStep, the \metaWithArchive{} improves reward by 0.185 on hard tasks but only 0.015 on easy tasks, where \noStudy{} already scores 0.792 (Table~\ref{tab:strata-difficulty}). On BCP-G, \corpusToSkill{} improves all ten topics, and its lift rises from 0.224 with one gold document to 0.293 with four or more (Tables~\ref{tab:strata-bcp-topic} and~\ref{tab:strata-bcp-documents}). Every method obtains its largest AppWorld gain on two-application tasks (Table~\ref{tab:strata-appworld-scale}) and its largest Apex Agents domain gain in Law (Table~\ref{tab:strata-apex-domain}). These patterns do not define a universal notion of task difficulty.

\textbf{Artifacts can sometimes misdirect the solver.} Appendix~\ref{app:test-trace-examples} contrasts two cases. In DABStep, an artifact supplies missing information about a fraud boundary that the solver uses correctly. In Apex Agents, an incomplete artifact directs the solver toward general pricing documents and away from two task-specific files containing the required inputs. Studying can therefore actively mislead the solver, for example by narrowing its search or hiding relevant evidence.

%% file: sections/discussion.tex
\section{Discussion and Limitations}
\label{sec:discussion}

Our results provide evidence that task-agnostic, environment-conditioned studying is generally beneficial: 21 of 24 method--benchmark comparisons improve over \noStudy{}. \emph{Dynamically} selecting how to study is robust but not universally optimal. A \metaAgent{} variant leads on five of six benchmarks, while the specialized \corpusToSkill{} pipeline is strongest on BCP-G. A promising direction is therefore to improve how the agent selects a studying strategy. A broader procedure archive or demonstrations of successful environment--strategy matches could improve selection in context. Across repeated deployments, downstream evaluation feedback could instead be used to update the \metaAgent{}'s strategy-selection policy -- via the weights or harness -- turning experience across environments into expertise, in a form of meta-learning.

Sustained study-budget scaling appears only on Harvey LAB and only for the \metaAgent{} methods. For \preping{}, this differs from the original study, which reports steady gains with additional synthetic tasks on AppWorld~\citep{preping}. Since our sweep covers different benchmarks, we note that the returns to additional synthetic practice may be environment-dependent. For the \metaAgent{}, the budget is stated in the prompt, but effective scaling requires more than budget awareness: the model must track its consumption and plan differently as the allowance changes. Current models may struggle with both, so a larger budget may extend the same strategy rather than induce a qualitatively different one. Our experiment therefore measures how the current policy responds to more budget, not what a studier with calibrated budget tracking and planning could achieve.

Studying has a clearer effect on task-time scaling, producing reusable structure that can be amortized across future tasks. One studied rollout reaches scores that require two to eight \noStudy{} rollouts under oracle selection. This benefit is not guaranteed, however, and our qualitative examples show that study artifacts can sometimes misdirect the solver in detrimental ways.

Our conclusions in this work are limited by scope and measurement. The agentic studiers and frozen solver use models from the same family and share the Claude Code harness, so we cannot establish whether the resulting artifacts transfer across solvers or provide solver-independent improvements. We evaluate only one workflow archive and do not compare the \metaAgent{} against a non-agentic classifier that selects one fixed workflow from the same archive based on an environment description. We therefore cannot pinpoint the value of agentic decision-making and workflow composition, the contribution of individual archive entries, or the effect of expanding the archive. Although our benchmarks span different corpus sizes, tool surfaces, and domains, six benchmarks cannot cover every axis along which agent environments vary. Finally, dollar cost provides a common currency for strategies that use different models, and sometimes multiple models, but collapses heterogeneous computation into a single provider-dependent proxy. A model-independent measure such as FLOPs would avoid this dependence but is difficult to obtain consistently, particularly for closed-source models.

%% file: sections/conclusion.tex
\section{Conclusion}
\label{sec:conclusion}
We study task-agnostic environment preparation, in which systems inspect an unfamiliar environment and construct artifacts for a frozen solver before the downstream task distribution is known. We compare fixed studying strategies with meta-agent variants that study freely or compose workflows from an archive. Across six benchmarks, a \metaAgent{} variant attains the highest Avg@3 reward on five, while \corpusToSkill{} is strongest on BCP-G; archive access produces occasional large gains but not uniform improvements. Additional study budget does not reliably improve reward, but studied artifacts reduce the test-time sampling needed to reach a given score and can sometimes misdirect the solver.

\paragraph{Responsible Use}
\label{sec:responsible}
Task-agnostic preparation may reduce the expert effort required to adapt agents to specialized environments. Studying systems also explore environments before target tasks are known and persist what they learn for later use, creating four risks. \textbf{(1) Exploration side effects.} Study may invoke costly or irreversible actions and should use sandboxes, least-privilege access, and approval gates. \textbf{(2) Sensitive artifacts.} Artifacts may preserve private data, environment structure, or credentials; they should inherit source access controls and retention policies. \textbf{(3) Misleading guidance.} Incomplete or stale artifacts can harm downstream performance, as our qualitative analysis illustrates. Provenance and human review can mitigate this risk. \textbf{(4) Archive misuse.} Reusable workflows can facilitate exploration in harmful environments. Archive entries should therefore specify permitted scopes and side-effect assumptions, and their execution should be logged.

%% file: sections/appendix_benchmarks.tex
\subsection{Benchmark and Environment Details}
\label{app:benchmark-surfaces}

Figure~\ref{fig:spectrum} and Table~\ref{tab:bench} provide complementary views of the counts and types of the files and tools in each of the benchmarks we investigate.

\input{figures/fig_spectrum.tex}

\begin{table}[h]
\centering
\footnotesize
\begin{tabular}{@{}p{0.14\textwidth}p{0.36\textwidth}p{0.43\textwidth}@{}}
\toprule
Benchmark & Benchmark-specific tools & Files exposed to the agent \\
\midrule
BCP-G~\citep{browsecompplus} & None & Approximately 100{,}195 text files, consistently observed across 17 traces \\
OfficeQA~\citep{officeqa} & None & 697 Treasury Bulletin text files \\
Harvey LAB~\citep{harveylab} & 1 CLI (\texttt{labread}) for binary documents & 9{,}288 documents in 266 matter folders (DOCX, XLSX, PPTX, EML) \\
DABStep~\citep{dabstep} & None (files require Python access) & 7: \texttt{payments.csv}, \texttt{fees.json}, \texttt{manual.md}, and 4 references, jointly comprising task data, answer-bearing references, and protocol instructions \\
Apex Agents~\citep{apexagents} & 9 MCP servers and 8 distinct observed tool entry points & Per-world file-system median $\sim$40, range 4--170, plus \texttt{/.apps\_data}, estimated across 20 worlds \\
AppWorld~\citep{appworld} & At least 105 distinct MCP tools across 8 apps in the union of 150 traces. The full API surface is larger & No mounted file corpus. State is stored in app databases, with files accessed through the virtual \texttt{file\_system} app \\
\bottomrule
\end{tabular}
\caption{Files and benchmark-specific tools available in each benchmark environment.}
\label{tab:bench}
\end{table}

%% file: figures/fig_spectrum.tex
\begin{figure}[h]
\centering
\begin{tikzpicture}[
  x=0.93cm,
  y=0.6cm,
  font=\scriptsize,
  point/.style={circle,draw=black,fill=green!35,inner sep=1.8pt},
  label/.style={align=center,fill=white,inner sep=1.5pt}
]
  \draw[-{Latex[length=2mm]}, thick] (0,0) -- (11.55,0);
  \node[below=8mm,align=center] at (5.75,0) {files exposed to agent (log scale)};
  \draw[-{Latex[length=2mm]}, thick] (0,0) -- (0,6.65);
  \node[rotate=90,align=center] at (-1.35,3.3) {benchmark-specific tools (log scale)};

  \foreach \x/\lab in {0/0,2.29/10,4.41/100,6.60/{1K},8.80/{10K},11.00/{100K}} {
    \draw (\x,0.08) -- (\x,-0.08) node[below=2pt] {\lab};
  }
  \foreach \y/\lab in {0/0,0.90/1,3.12/10,6.01/100} {
    \draw (0.08,\y) -- (-0.08,\y) node[left=2pt] {\lab};
  }

  \node[point] (dab) at (1.99,0) {};
  \node[label,above=5pt of dab] {DABStep\\7 files};
  \node[point] (office) at (6.26,0) {};
  \node[label,above=5pt of office] {OfficeQA\\697 files};
  \node[point] (harvey) at (8.73,0.90) {};
  \node[label,above=5pt of harvey] {Harvey LAB\\1 tool, 9{,}288 files};
  \node[point] (bcp) at (11.00,0) {};
  \node[label,above left=5pt and 0pt of bcp] {BCP-G\\$\sim$100K files};

  \node[point] (appworld) at (0,6.08) {};
  \node[label,right=5pt of appworld] {AppWorld\\$\geq$105 tools};

  \draw[thick] (1.54,2.86) -- (4.91,2.86);
  \draw[thick] (1.54,2.70) -- (1.54,3.02);
  \draw[thick] (4.91,2.70) -- (4.91,3.02);
  \node[point] (apex) at (3.55,2.86) {};
  \node[label,above=5pt of apex] {Apex Agents\\8 tool entry points};
\end{tikzpicture}
\caption{\textbf{File and tool counts across benchmark environments.} Axes show $\log_{10}(1+n)$ so environments with 0 benchmark-specific tools or real files remain visible. Apex's horizontal interval is its per-world file range (4--170, median $\sim$40). AppWorld exposes at least 105 distinct tools across observed traces, while its state lives in app databases rather than real files. Generic shell and Python operations are not counted as benchmark-specific tools.}
\label{fig:spectrum}
\end{figure}
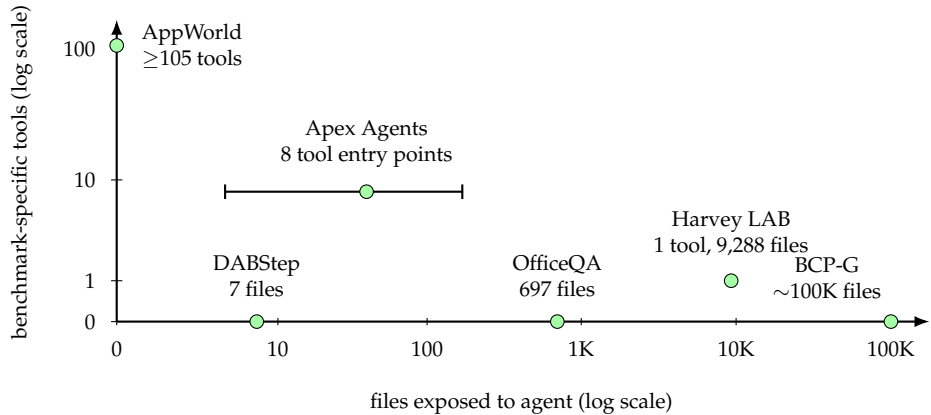

%% file: sections/appendix_results.tex
\subsection{Additional Results}
\label{app:additional-results}
\label{app:cost-by-benchmark}
\label{app:apex-domain-lift}

This section reports additional performance and cost results that accompany the main analysis.

\begin{table}[h!]
\centering
\small
\begingroup
\setlength{\tabcolsep}{5.5pt}
\renewcommand{\arraystretch}{1.16}
\resizebox{\textwidth}{!}{%
\begin{tabular}{@{}lllllll@{}}
\toprule
\textbf{Method} & \multicolumn{1}{c}{\textbf{BCP-G}} & \multicolumn{1}{c}{\textbf{OfficeQA}} & \multicolumn{1}{c}{\textbf{Harvey LAB}} & \multicolumn{1}{c}{\textbf{DABStep}} & \multicolumn{1}{c}{\textbf{Apex Agents}} & \multicolumn{1}{c}{\textbf{AppWorld}} \\
\midrule
\noStudy{}                & $0.540 \pm 0.011$ & $0.272 \pm 0.004$ & $0.239 \pm 0.030$ & $0.125 \pm 0.008$ & $0.236 \pm 0.009$ & $0.197 \pm 0.007$ \\
\addlinespace[2pt]
\rowcolor{tablestripe}
\preping{}                & $0.645 \pm 0.070$ & $0.246 \pm 0.007$ & $0.243 \pm 0.052$ & $0.135 \pm 0.011$ & $0.290 \pm 0.017$ & $0.262 \pm 0.067$ \\
\corpusToSkill{}          & $0.783 \pm 0.053$ & $0.450 \pm 0.066$ & $0.309 \pm 0.011$ & $0.129 \pm 0.001$ & $0.279 \pm 0.016$ & $0.212 \pm 0.011$ \\
\addlinespace[2pt]
\rowcolor{tablestripe}
\metaNoArchive{}          & $0.620 \pm 0.036$ & $0.248 \pm 0.007$ & $0.258 \pm 0.007$ & $0.139 \pm 0.004$ & $0.248 \pm 0.014$ & $0.206 \pm 0.023$ \\
\metaWithArchive{}        & $0.751 \pm 0.122$ & $0.255 \pm 0.014$ & $0.265 \pm 0.029$ & $0.132 \pm 0.016$ & $0.256 \pm 0.027$ & $0.184 \pm 0.012$ \\
\bottomrule
\end{tabular}}
\endgroup
\caption{\textbf{Average inference cost per rollout.} Values are mean $\pm$ sample SD in USD. Studying-method statistics are across three independent artifact iterations. \noStudy{} statistics are across 10 full task-time repetitions. Unpriced rollouts are excluded rather than imputed.}
\label{tab:inference-cost-by-benchmark}
\end{table}

\begin{table}[h!]
\centering
\small
\begingroup
\setlength{\tabcolsep}{5.5pt}
\renewcommand{\arraystretch}{1.16}
\resizebox{\textwidth}{!}{%
\begin{tabular}{@{}lllllll@{}}
\toprule
\textbf{Method} & \multicolumn{1}{c}{\textbf{BCP-G}} & \multicolumn{1}{c}{\textbf{OfficeQA}} & \multicolumn{1}{c}{\textbf{Harvey LAB}} & \multicolumn{1}{c}{\textbf{DABStep}} & \multicolumn{1}{c}{\textbf{Apex Agents}} & \multicolumn{1}{c}{\textbf{AppWorld}} \\
\midrule
\noStudy{}                & -- & -- & -- & -- & -- & -- \\
\addlinespace[2pt]
\rowcolor{tablestripe}
\preping{}                & $24.35 \pm 3.90$ & $15.26 \pm 10.43$ & $17.57 \pm 12.37$ & $11.71 \pm 8.26$ & $681.79 \pm 7.63$ & $23.97 \pm 8.61$ \\
\corpusToSkill{}          & $127.65 \pm 110.56$ & $1.42 \pm 0.02$ & $1.15 \pm 0.02$ & $0.01 \pm 0.00$ & $17.49 \pm 0.02^{\ddagger}$ & $1.04 \pm 0.05$ \\
\addlinespace[2pt]
\rowcolor{tablestripe}
\metaNoArchive{}          & $0.68 \pm 0.19$ & $0.76 \pm 0.09$ & $7.36 \pm 4.10$ & $0.59 \pm 0.09$ & $66.96 \pm 8.37$ & $2.18 \pm 0.43$ \\
\metaWithArchive{}        & $97.87 \pm 112.73$ & $15.99 \pm 9.67$ & $31.21 \pm 22.94$ & $19.47 \pm 0.65$ & $584.21 \pm 198.78$ & $13.83 \pm 19.72$ \\
\bottomrule
\end{tabular}}
\endgroup
\caption{\textbf{Average pre-task cost.} Values are mean $\pm$ sample SD in USD. Statistics are across three independent artifact-construction iterations. \noStudy{} performs no pre-task computation and is shown as dashes. The Apex Agents \corpusToSkill{} cell, marked $\ddagger$, uses the two metered artifact-construction iterations because the original iteration did not retain cost telemetry.}
\label{tab:pretask-cost-by-benchmark}
\end{table}

\clearpage
Figure~\ref{fig:apex-domain-lift} disaggregates the \metaWithArchive{} Avg@3 lift over \noStudy{} across individual Apex Agents worlds, grouped by domain.

\input{figures/fig_apex_domain_lift.tex}

\clearpage

%% file: figures/fig_apex_domain_lift.tex
\begin{figure}[h!]
\centering
\includegraphics[width=\textwidth]{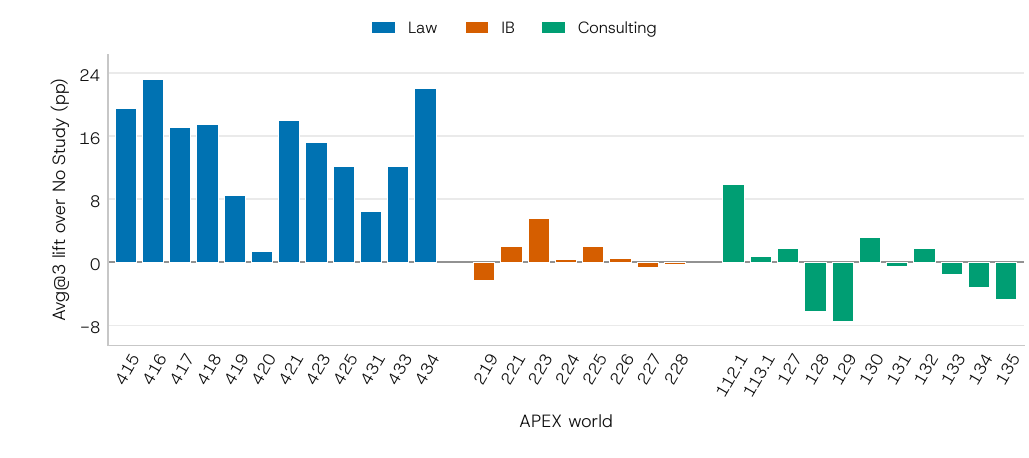}
\caption{\textbf{\metaWithArchive{} lift by APEX Agents world.} Each bar shows the signed percentage-point difference between \metaWithArchive{} Avg@3 and the \noStudy{} Avg@3 baseline for one APEX Agents world. Within each of three independent artifact iterations, the studied score initially averages the three evaluation rewards for each task and then macro-averages tasks. The plotted score averages those three artifact-iteration estimates. For \noStudy{}, we compute the three-repetition mean for each of the $\binom{10}{3}$ subsets and average those subset estimates. This point estimate is algebraically equal to the 10-repetition mean. Bar color denotes domain. World scores macro-average the tasks in that world.}
\label{fig:apex-domain-lift}
\end{figure}

%% file: sections/appendix_methods.tex
\subsection{Additional Implementation Details}
\subsubsection{Studying method configuration}
\label{app:implementation-details}

This section reports implementation choices not specified in the main text.

\paragraph{Common configuration.}
Agentic study runs use Claude Code with permission bypass enabled and web access disabled; otherwise, they retain provider defaults.

%

\paragraph{\preping{}.}
Synthetic tasks run through the corresponding benchmark's task-agent harness using a task-neutral environment projection. We use a feasibility threshold of 5/5 and treat completion scores of at least 4/5 as successful. Proposal and validation temperatures are 0.7 and 0, respectively. During construction and delivery, at most 40 playbook entries are retrieved using \texttt{text-embedding-3-small}. Individual practice rollouts are not dollar-capped. Only validator-approved feasible trajectories are reflected on and curated into the playbook.

\paragraph{\corpusToSkill{}.}
We use the published defaults except for benchmark-specific adjustments to hierarchy branching, document length, and tree compaction. We increase the branching ratio from 10 to 40 on OfficeQA and Harvey LAB. We use a 4,000-character document limit and compact tree representation on DABStep, OfficeQA, and Harvey LAB; the remaining benchmarks retain the default 8,000-character limit and non-compact representation. 

\paragraph{Artifact delivery boundary.}
\noStudy{} makes no environment or prompt change. Each studying method exposes only its frozen output through its native interface: retrieved playbook entries for \preping{}, the compiled skill hierarchy and document accessor for \corpusToSkill{}, and mounted artifacts for the two \metaAgent{} variants. The original target instruction is preserved in every arm.

\input{sections/appendix_prompts.tex}

%% file: sections/appendix_prompts.tex
\subsubsection{Representative implementation prompts}
\label{app:representative-prompts}
The listings below show one representative prompt for each shared method component needed to interpret the intervention. Benchmark-specific environment clauses, proposer and validator variants, and submission suffixes are omitted here. The implementation and released run records contain the complete rendered prompts. Punctuation and formatting are normalized to the manuscript style.

\paragraph{\preping{} task proposer.}
\begin{Verbatim}[breaklines=true,breakanywhere=true,fontsize=\scriptsize]
You are a synthetic task generator for an AI agent to construct the memory in the environment.

Your goal is to generate diverse, realistic, and challenging task instructions that an AI assistant would need to complete. These tasks should reflect real-world user scenarios.

## API Documentation
{api_docs}

## Guidelines
1. **Feasibility**: Generate tasks that are actually executable with the given APIs
2. **Naturalness**: Write as a real user would phrase them, not as technical API calls
3. **Verifiability**: Instructions must be precise enough to allow for an objective judgment of success or failure.
   - The agent will submit its answer as a single text string. Success is judged by comparing that string to the expected value.
   - For this to be verifiable, each QUERY must require **exactly one type of answer** (e.g. one number, or one name/title, or one list.)
   - Do not ask for two or more distinct values in one task (e.g. both a count and a list, or both title and artist). Otherwise the expected format of the submitted text is ambiguous and cannot be reliably checked.
4. **Entity Use**:
   - Prefer realistic, natural task phrasing over rigid placeholder patterns.
   - If **environment information is provided**, treat it as optional grounding support:
     - You may use discovered names, titles, folders, contacts, and other entities when they help make a task concrete.
     - You do not need to force those entities into every task.
   - If **environment information is empty or missing**, you may use generic user-centric references such as:
     - "my favorite playlist", "my song library", "my playlists", "my album library", "artists I follow"

## Task Category Guidelines
Generate tasks across the following categories:

### 1. QUERY Tasks (Information Retrieval)
Tasks that require finding and returning specific information. The answer should be objective and verifiable (number, name, comma-separated list, yes/no).
Include concrete details with specific output format such as numbers, dates, names, locations, thresholds, item lists, etc.
Avoid vague terms such as 'check', 'review', 'show' or 'ensure' unless they are accompanied by measurable criteria.

### 2. ACTION Tasks (State Modification)
Tasks that require modifying the state of one or more apps. No explicit answer required - success is determined by state changes.

{environment_info_section}
{task_history_section}

## Output Format
Return a JSON array of task instructions:
```json
[
    {{
        "category": "QUERY|ACTION",
        "involved_apps": ["app1", "app2"],
        "involved_apis": ["app.api1", "app.api2"],
        "instruction": "Natural language task instruction"
    }}
]
```

Generate {num_tasks} diverse task instructions:
\end{Verbatim}

\paragraph{\preping{} trajectory validator.}
\begin{Verbatim}[breaklines=true,breakanywhere=true,fontsize=\scriptsize]
You are an expert evaluator of agent task execution.

Evaluate the following Task and Trajectory using a 5-point Likert scale for each criterion.

## Evaluation Criteria

### 1. Task Feasibility
Evaluate whether the task is executable and the entities are grounded in the current environment.
- Do not judge feasibility by API availability alone.
- Verify that every required target entity in the instruction (person, contact, request, file, thread, project, etc.) is present and can be uniquely identified from the trajectory evidence.
- If the entity cannot be identified after reasonable paginated search using relevant APIs, the task is infeasible.

- **5 (Excellent)**: All required entities/constraints are explicitly present and uniquely identified. Clearly executable as stated.
- **4 (Good)**: Entities are present with minor ambiguity resolvable by a simple lookup. No core entity missing.
- **3 (Acceptable)**: Plausible, but key grounding is uncertain or search is insufficient.
- **2 (Poor)**: Likely infeasible. Required entity not found or not uniquely identifiable after reasonable search.
- **1 (Unacceptable)**: Infeasible/contradictory or relies on non-existent entities/conditions.

### 2. Task Completion
Judge whether the task instruction was successfully completed.

- Evaluate all task instructions are satisfied based on the trajectory, including all constraints.
- Treat a call to `apis.supervisor.complete_task(...)` as a required completion step, not as evidence that the task was successful correctly.
- If the task asked for action, do not treat the task as successful unless the trajectory includes a call to `apis.supervisor.complete_task()` with no answer argument (or with `None`).
- If the task asked for information, success requires a call to `apis.supervisor.complete_task(answer=<answer>)`, and the answer should contain only the exact requested value: a number (int or float), a direct string value, a plain comma-separated list, or `yes`/`no`.
- Do not treat verbose answers, action summaries, explanations, full sentences, or answers with extra words, units, or symbols as correct unless explicitly requested. e.g. `10` is acceptable, but `The answer is 10 songs.` is not. Do not require an answer for action-only tasks.

- **5 (Excellent)**: Every requirement AND every constraint is satisfied, and the trajectory shows clear completion for each required outcome.
- **4 (Good)**: All requirements/constraints appear satisfied with minor ambiguity, and nothing important is missing.
- **3 (Acceptable)**: Some progress, but at least one requirement/constraint is missing, ambiguous, or only asserted without support in the trajectory steps (surface-level success). This includes cases where an object/ID exists but correctness constraints (privacy, filters, counts, formatting, attachment actually uploaded, etc.) are not confirmed.
- **2 (Poor)**: Minor progress only. The main outcome is not achieved or an error/failed response prevents completion.
- **1 (Unacceptable)**: No meaningful progress toward the task goal.

## Task Instruction
{task_instruction}

## Trajectory
{trajectory}

Respond in JSON format:
```json
{{
  "feasibility_reason": "1-2 sentence explanation for feasibility score",
  "feasibility_score": 1-5,

  "task_completion_reason": "1-2 sentence explanation for task completion score",
  "task_completion_score": 1-5
}}
```
\end{Verbatim}

\paragraph{\preping{} reflector.}
\begin{Verbatim}[breaklines=true,breakanywhere=true,fontsize=\scriptsize]
You are an expert agent and educator. Your job is to diagnose the current trajectory: identify what went wrong (or could be better), API usage, and ground truth when applicable.

Instructions:
- Carefully analyze the model's reasoning trace to identify where it went wrong
- Take the environment feedback into account, comparing the predicted answer with the (optional) ground truth to understand the gap
- Identify specific conceptual errors, calculation mistakes, or misapplied strategies
- Provide actionable insights that could help the model avoid this mistake in the future
- Identify root causes: wrong source of truth, bad filters (timeframe/direction/identity), formatting issues, or missing authentication and how to correct them.
- Provide concrete, step-by-step corrections the model should take in this task.
- Be specific about what the model should have done differently
- You will receive bulletpoints that are part of playbook that's used by the generator to answer the question.
- You need to analyze these bulletpoints, and give the tag for each bulletpoint, tag can be ['helpful', 'harmful', 'neutral'] (for the generator to generate the correct answer)
- Explicitly curate from the environment feedback the output format/schema of APIs used when unclear or mismatched with expectations (e.g., apis.blah.show_contents() returns a list of content_ids (strings), not content objects)

Inputs:

Task Instruction:
{{task_description}}

{{ground_truth_result}}

{{ground_truth_code}}
{{unit_test_results}}
PrePing playbook (playbook that's used by model for code generation):
PLAYBOOK_START
{{playbook}}
PLAYBOOK_END

Agent-Environment Trajectory (including reasonings actions, and final observation):
{{trajectory}}

Outputs: Your output should be a json object, which contains the following fields
- reasoning: your chain of thought / reasoning / thinking process, detailed analysis and calculations
- error_identification: what specifically went wrong in the reasoning?
- root_cause_analysis: why did this error occur? What concept was misunderstood?
- correct_approach: what should the model have done instead?
- key_insight: what strategy, formula, or principle should be remembered to avoid this error?
- bullet_tags: a dictionary mapping each bullet_id (the {section}-{number} prefix shown in the playbook) to its tag ('helpful', 'harmful', or 'neutral')
Answer in this exact JSON format:

{
"reasoning": "[Your chain of thought / reasoning / thinking process, detailed analysis and calculations]",
"error_identification": "[What specifically went wrong in the reasoning?]",
"root_cause_analysis": "[Why did this error occur? What concept was misunderstood?]",
"correct_approach": "[What should the model have done instead?]",
"key_insight": "[What strategy, formula, or principle should be remembered to avoid this error?]",
"bullet_tags": {"bullet_id_1": "helpful", "bullet_id_2": "harmful", "bullet_id_3": "neutral"}
}
\end{Verbatim}

\paragraph{\preping{} curator.}
\begin{Verbatim}[breaklines=true,breakanywhere=true,fontsize=\scriptsize]
You are a master curator of knowledge. Your job is to identify what new insights should be added to an existing playbook based on a reflection from a previous attempt.

Context:
- The playbook you created will be used to help answering similar questions.

Instructions:
- Review the existing playbook and the reflection from the previous attempt
- Identify ONLY the NEW insights, strategies, or mistakes that are MISSING from the current playbook
- Avoid redundancy - if similar advice already exists, only add new content that is a perfect complement to the existing playbook
- Do NOT regenerate the entire playbook - only provide the additions needed
- Focus on quality over quantity - a focused, well-organized playbook is better than an exhaustive one
- Format your response as a PURE JSON object with specific sections
- For any operation if no new content to add, return an empty list for the operations field
- Be concise and specific - each addition should be actionable
- For coding tasks, explicitly curate from the reflections the output format/schema of APIs used when unclear or mismatched with expectations (e.g., apis.blah.show_contents() returns a list of content_ids (strings), not content objects)

Task Instruction:
{question_context}

Current Playbook:
{current_playbook}

Agent-Environment Trajectory (actions and outputs from the attempt):
{trajectory}

Current Reflections (principles and strategies that helped to achieve current task):
{guidebook}

Your Task:
Output ONLY a valid JSON object with these exact fields:
- reasoning: your chain of thought / reasoning / thinking process, detailed analysis and calculations
- operations: a list of operations to be performed on the playbook
- type: the type of operation to be performed
- section: the section to add the bullet to (one of: strategies, code_snippets, pitfalls, apis)
- content: the new content of the bullet

Available Operations:
1. ADD: Create new bullet points with fresh IDs
- section: the section to add the new bullet to
- content: the new content of the bullet. Note: no need to include the bullet_id in the content like '[ctx-00263] helpful=1 harmful=0 ::', the bullet_id will be added by the system.

RESPONSE FORMAT - Output ONLY this JSON structure (no markdown, no code blocks):
{{
"reasoning": "...",
"operations": [
    {{ "type": "ADD", "section": "...", "content": "..." }}
]
}}
\end{Verbatim}

\paragraph{Unaided exploratory-study prompt, represented by BCP-G.}
\begin{Verbatim}[breaklines=true,breakanywhere=true,fontsize=\scriptsize]
ENVIRONMENT : BCP-Grep study. The material you are studying (the "corpus" referred to below) is a fixed collection of plaintext documents under `/workspace/corpus/`. Each `<docid>.txt` file starts with its `DOCID` and `URL`, followed by the complete document text. Use Claude Code's native `Grep` across `/workspace/corpus`, then `Read` matching files. Use `Glob` when filename discovery is useful. There is no web access. You may write ONLY under `/logs/artifacts/cheatsheet/`.

You are studying for a test that you have no prior information about other than that it will be based on the environment you have been placed in : the documents it contains, and the tools available to you for finding and reading them. You are not allowed to search the web at any point or to go outside this environment. You are given a limited budget in which you may explore as you wish: inspect the corpus to learn where different kinds of information live and exercise the available tools to learn what they do and how they behave. Your goal is to leave behind a minimal set of artifacts in the /logs/artifacts/cheatsheet/ directory that inform a future, budget-constrained version of yourself as much as possible without overwhelming it : maximize useful signal per token, and add nothing that is merely noise. A test-time agent will have the same access you do, so its budget is wasted on exactly two things: finding where relevant information lives in an unfamiliar corpus, and figuring out how to operate unfamiliar tools correctly. Spend your study budget mapping those two things and little else : where to look for a given kind of information, and how to use each non-obvious tool (its purpose, the few calls that matter, and any quirks or pitfalls you actually hit). Prefer pointers and short operating notes over copied content: say where to look and how to act rather than reproducing material, which can go stale and invites blind reliance. Be ruthless about clutter: do not record anything obvious from a document's title or a tool's name, anything a test-time agent could rediscover in a step or two, or general knowledge you already possess : if a line does not save meaningful test-time budget, delete it. At test time you will be given read access to whatever you created in /logs/artifacts/cheatsheet/ along with the full environment, but you will be on a limited inference budget, so the quality, brevity, and accessibility of what you leave behind may greatly affect your test-time performance. Proceed however you like, but the only write access you have is to the /logs/artifacts/cheatsheet/ directory, and you may explore the corpus through its native filesystem tools and the /logs/artifacts/cheatsheet/ directory.

:

GROUNDING REQUIREMENT: every fact, entity, claim, entry, table row, Q&A pair, map node, or note recorded in an artifact MUST cite the exact numeric document ID and file path, for example `[5412] /workspace/corpus/5412.txt`. A future agent must be able to open the cited file directly with `Read` instead of repeating discovery. For tool-operating notes, name the native tool, minimal call pattern, path, and any observed quirk or pitfall.
\end{Verbatim}

\paragraph{\corpusToSkill{} downstream navigation prompt.}
\begin{Verbatim}[breaklines=true,breakanywhere=true,fontsize=\scriptsize]
You are navigating a closed, precompiled corpus through Claude Code skills and one exact-ID document tool.

For every task:

1. Invoke at least two plausible top-level skills with the Skill tool before settling on a branch.
2. Descend through the selected skills by reading their `SKILL.md` and nested `INDEX.md` files with Read. Use Glob to discover relevant index files and Grep only within the skill/index tree. Consult `entity_index.json` when named entities can disambiguate the route. If a branch is thin or unhelpful, cross-jump to a related or second plausible skill.
3. Select exact document IDs from leaf indexes before calling `get_document`. The document tool does not search or list the corpus.
4. Make at least one successful `get_document` call. Treat retrieved full documents, rather than skill summaries, index summaries, or entity mappings, as factual evidence for the answer.

Do not use corpus-search tools or open-web tools. Do not read, grep, parse, or run scripts against `documents.json`. Access full documents only through `get_document` with IDs found in leaf indexes.
\end{Verbatim}

\paragraph{\metaWithArchive{} Prompt}
\begin{Verbatim}[breaklines=true,breakanywhere=true,fontsize=\scriptsize]
# Preparation goal

Prepare one frozen, task-blind bundle that gives the later target solver the best chance of solving
unknown tasks from this environment group. You may inspect only the neutral environment projection.
real target instructions, verifier code, metric details, answer-bearing files, and real baseline
traces are unavailable and must remain unavailable.

You are the controller. Every synthetic probe and every eventual target run uses the pinned
target solver through the real harness. The elapsed deadline began before this session invocation.
resuming never resets it. The gateway is the authority for the deadline, active work, costs, and
committed final state. If the main deadline fires, all exploration/workflow/probe tools are disabled
and active jobs are canceled. A short hard-bounded grace permits only working-note updates and
prefix/bundle consolidation from workflow jobs that completed before the deadline. It does not
permit new artifacts, probes, or workflow execution. Expiry commits `timeout_forced_null`.

## Durable work

`working-notes.md` in durable scratch is your authoritative diagnosis and decision record. Update it:

- after your initial diagnosis,
- after every meaningful probe or workflow result,
- before entering a long `await_jobs` call,
- whenever you change the candidate stack, and
- immediately before finalization.

Keep draft prefix text, your ordinary artifacts, and workflow checkpoints in durable scratch. A
session may be compacted or interrupted, so do not leave the only copy of a decision in conversation
history or transient filesystem state.

Use `update_working_notes` so each update is fsynced and recorded in the append-only ledger.

## Workflows and probes

The control skills describe Cheatsheet, PREPING, Corpus2Skill, and synthetic probe invocation without
performance or routing advice. You may invoke, configure, combine, compare, or decline them. Select
at most one successful output from each workflow family. PREPING delivery is one complete static
playbook shared by the group. It must never retrieve against a real target instruction. Controller-authored
files are ordinary inert artifacts and never new Claude Code skills.

Gateway submissions return durable job IDs immediately. When work is outstanding, call
`await_jobs(job_ids, return_when)` once and let it block. Do not implement sleeps, polling loops,
repeated status prompts, or conversational check-ins. The wait wakes for a requested terminal job,
the 30-minute warning, the final deadline, or infrastructure requiring attention. Before a long
wait, update working notes.

## Final bundle

The admitted `/meta_archive` layout is fixed: `manifest.toml`, `prefix.md`, canonical workflow skills
under `skills/`, workflow artifacts under `artifacts/{cheatsheet,preping,corpus2skill}/`, your inert
artifacts under `artifacts/meta/`, and declared additive tools/data under `tools/`. Seed control skills
never transfer. Native solver tools and filesystem access remain available. Canonical prompt blocks
describe actual validated knobs and allocated deliverables. You may weave, reorder, shorten, extend,
or decline them, but every interface named in the final prefix must be truthful.

Call `finalize_bundle` to validate and lock the static bundle, cancel unselected work, and end
preparation. You may instead call `choose_null`. A deadline or unrecoverable crash forces null. All
null states receive the unmodified baseline handoff. Never write or infer a real task instruction.
\end{Verbatim}

%% file: sections/appendix_interpretability.tex
\section{Study Time Interpretability}
\label{app:pretask-interpretability}

This section details our methods for analyzing how the studying systems explored each environment, which behaviors they exhibited, and what they produced. 

\subsection{Environment Exploration Analysis}
\label{app:study-coverage}

We measure environment exploration using the number of unique files read and benchmark-specific tools invoked during study. Figures~\ref{fig:study-files-vs-score} and~\ref{fig:study-tools-vs-score} relate these quantities to downstream reward for each benchmark, method, and artifact iteration.

\input{figures/fig_study_files_vs_score.tex}
\input{figures/fig_study_tools_vs_score.tex}

\paragraph{Measurement.} For trace-based methods, we count a file when its contents appear in the study trace through a direct read or an attributed search result. Directory listings and filename mentions do not count. For \corpusToSkill{}, which has no interactive trace, we instead count the documents processed by its compiler. We count each benchmark-specific tool invoked at least once, including failed calls. Shell and generic file operations are excluded.

\paragraph{Scope.} Counts are aggregated over the available study traces for each benchmark, method, and artifact iteration. Apex Agents additionally aggregates across its 31 worlds. DABStep is omitted from the file analysis because every method reads all seven files, and AppWorld because it exposes no real files. Trace-derived file counts are lower bounds because internal access and truncated results may not expose every file read.

\paragraph{Results.} We fit a Theil--Sen line to the points for each benchmark. These fits are descriptive: exploration and method vary together, so they do not isolate the effect of reading an additional file or invoking an additional tool. File-count slopes are nearly flat for Harvey LAB and Apex Agents, negative for OfficeQA, and positive for BCP-G. The stronger OfficeQA and BCP-G associations are driven largely by \corpusToSkill{}, which processes the entire corpus and therefore lies far from the trace-based methods on the file axis. Tool-count slopes are also nearly flat for Apex Agents and AppWorld. Harvey LAB is the exception: both \metaAgent{} variants invoke \texttt{labread} and outperform methods that do not, although tool use and method identity are confounded. Overall, exploration breadth may matter, but it does not by itself determine downstream performance.

\subsection{Composition of Archive-Assisted Artifacts}
\label{app:archive-usage}

\metaWithArchive{} can invoke three studying procedures: the open-ended procedure used by \metaNoArchive{}, \preping{}, and \corpusToSkill{}. It can inspect their outputs and combine the resulting artifacts for the solver. Table~\ref{tab:archive-workflow-usage} reports how often artifacts from each procedure appeared in the final artifact set.

\begin{table}[h]
\centering
\small
\resizebox{\textwidth}{!}{%
\begin{tabular}{@{}lcccccc@{}}
\toprule
Artifact source & BCP-G & OfficeQA & Harvey LAB & DABStep & Apex Agents & AppWorld \\
\midrule
Open-ended study & 3/3 & 3/3 & 3/3 & 3/3 & 68/93 & 2/3 \\
\preping{}        & 3/3 & 3/3 & 2/3 & 3/3 & 34/93 & 1/3 \\
\corpusToSkill{}  & 0/3 & 0/3 & 2/3$^{\dagger}$ & 0/3 & 86/93 & 3/3 \\
\bottomrule
\end{tabular}}
\caption{Sources of artifacts included by \metaWithArchive{}. Each cell reports the number of final artifact sets containing output from that procedure. Multiple sources may appear in one artifact set. Iterations are pooled. Apex Agents is additionally pooled across worlds. $\dagger$~\corpusToSkill{} was unavailable in the first Harvey LAB iteration and was included in both iterations in which it was available.}
\label{tab:archive-workflow-usage}
\end{table}

\paragraph{Results.} Artifact composition varies across environments. Open-ended study and \preping{} appear in every BCP-G, OfficeQA, and DABStep artifact set, whereas \corpusToSkill{} appears in none. In contrast, \corpusToSkill{} appears in every AppWorld artifact set, both available Harvey LAB sets, and 86 of 93 Apex Agents sets. However, the same procedures were generally invoked across environments, and an output may be absent because its procedure failed or because \metaWithArchive{} did not retain it. The table therefore describes final artifact composition rather than invocation decisions. Composition is also imperfect: \metaWithArchive{} omits \corpusToSkill{} on BCP-G even though it is the strongest standalone method there.

\subsection{Taxonomies of \metaAgent{} Study Behavior and Artifacts}
\label{app:meta-taxonomies}

We analyze the \metaAgent{} variants in two ways: the behaviors visible in their study traces and the content of their final artifacts.

\paragraph{Study behavior taxonomy.} We define nine multi-label categories for actions visible in a study trace:

\begin{itemize}
  \item \textbf{Environment inventory before content.} Enumerating the environment's whole surface (file tree, corpus size, app list) with the agent's own listing or counting calls before the first content-bearing read.
  \item \textbf{Interface probing.} Issuing calls whose purpose is to learn the interface itself, including help probes, format tests, trying a tool to see whether it exists, or sweeping logins to confirm an authentication pattern, and recording the lesson.
  \item \textbf{Documentation verification.} Explicitly testing a claim made by shipped documentation or metadata against observed data and recording the contradiction or confirmation.
  \item \textbf{Structured sampling or hypothesis testing.} Sampling by named strata (per decade, ID range, app, or document type), or stating a hypothesis about the environment's organization and running a call to confirm or refute it, including bisecting a discovered boundary.
  \item \textbf{Recording unavailable resources.} Recording something as absent, empty, broken, or impossible so that the downstream agent does not spend time attempting to use it.
  \item \textbf{Budget-conscious extraction.} Making an explicit economizing decision: using cheap shell extraction before LLM reading, storing pointers instead of copied content, avoiding duplication of information already available to the solver, or timing operations to assess their cost.
  \item \textbf{Parallel sub-agent delegation.} Delegating exploration to parallel sub-agents with an explicit work partition and a specified output format or budget rule.
  \item \textbf{Artifact verification.} A dedicated pass checking claims destined for or already in the artifact against the environment: auditing cited paths, recomputing a recorded value, testing the artifact with mock questions, or a corrective edit after a re-check.
  \item \textbf{Test anticipation and rehearsal.} Predicting what the test will ask from a concrete task-bearing surface (inboxes, chat threads, an active sample task) and acting on the inference, including rehearsing a realistic task end-to-end.
\end{itemize}

\paragraph{Trace labeling.} Claude Sonnet~4.6 labels every analyzable trace at temperature 0 against the fixed category definitions. ``w/o'' denotes direct \metaNoArchive{} study, while ``w/'' denotes the open-ended study procedure invoked by \metaWithArchive{}. Each positive label requires a verbatim evidence quote, which we mechanically verify appears in the trace. 

\begin{table}[h]
\centering
\scriptsize
\resizebox{\textwidth}{!}{%
\begin{tabular}{@{}l*{12}{c}@{}}
\toprule
& \multicolumn{2}{c}{BCP-G} & \multicolumn{2}{c}{OfficeQA} & \multicolumn{2}{c}{Harvey LAB} & \multicolumn{2}{c}{DABStep} & \multicolumn{2}{c}{Apex Agents} & \multicolumn{2}{c}{AppWorld} \\
\cmidrule(lr){2-3}\cmidrule(lr){4-5}\cmidrule(lr){6-7}\cmidrule(lr){8-9}\cmidrule(lr){10-11}\cmidrule(lr){12-13}
Behavior & w/o & w/ & w/o & w/ & w/o & w/ & w/o & w/ & w/o & w/ & w/o & w/ \\
\midrule
Environment inventory before content & 3/3 & 3/3 & 3/3 & 3/3 & 3/3 & 3/3 & 3/3 & 3/3 & 93/93 & 73/73 & n/a & n/a \\
Interface probing                     & 3/3 & 3/3 & 2/3 & 2/3 & 3/3 & 3/3 & 0/3 & 0/3 & 84/93 & 68/73 & 3/3 & 1/2 \\
Documentation verification            & 2/3 & 0/3 & 0/3 & 0/3 & 1/3 & 3/3 & 2/3 & 3/3 & 45/93 & 43/73 & 0/3 & 0/2 \\
Structured sampling / hypothesis testing & 2/3 & 3/3 & 3/3 & 2/3 & 2/3 & 3/3 & 2/3 & 2/3 & 63/93 & 58/73 & 2/3 & 1/2 \\
Recording unavailable resources       & 3/3 & 3/3 & 3/3 & 3/3 & 3/3 & 3/3 & 3/3 & 3/3 & 89/93 & 70/73 & 2/3 & 2/2 \\
Budget-conscious extraction           & 3/3 & 3/3 & 3/3 & 3/3 & 3/3 & 3/3 & 3/3 & 3/3 & 91/93 & 71/73 & 3/3 & 2/2 \\
Parallel sub-agent delegation         & 0/3 & 0/3 & 0/3 & 0/3 & 3/3 & 2/3 & 0/3 & 0/3 & 14/93 & 10/73 & 0/3 & 0/2 \\
Artifact verification                 & 0/3 & 1/3 & 1/3 & 2/3 & 2/3 & 2/3 & 0/3 & 1/3 & 35/93 & 23/73 & 1/3 & 1/2 \\
Test anticipation and rehearsal       & 0/3 & 0/3 & 0/3 & 1/3 & 0/3 & 0/3 & 0/3 & 1/3 & 79/93 & 66/73 & 3/3 & 2/2 \\
\bottomrule
\end{tabular}}
\caption{Study behaviors observed in available trajectories. Each cell reports trajectories exhibiting the behavior out of all analyzable trajectories. AppWorld inventory is not applicable because its tool schemas are supplied rather than discovered through trace actions.}
\label{tab:study-action-taxonomy}
\end{table}

\paragraph{Behavior findings.} Environment inventory, recording unavailable resources, and budget-conscious extraction appear in nearly all applicable traces. Test anticipation is concentrated in Apex Agents and AppWorld, where histories and examples provide clues about likely downstream tasks. Parallel sub-agent delegation appears only in Harvey LAB and Apex Agents. The \metaAgent{} variants therefore share a common diagnostic core while adapting some exploration strategies to the environment.

\paragraph{Artifact taxonomy.} We define twelve multi-label categories for content appearing anywhere in a preparation's final artifact set:

\begin{itemize}
  \item \textbf{Navigation architecture.} Routing indexes, need-to-path or metric-to-source tables, file or app maps, or escalation recipes into deeper artifact tiers.
  \item \textbf{Method recipes.} Concrete commands, code snippets, or step-by-step procedures for recurring task shapes.
  \item \textbf{Environment constants.} Verified formats, units, formulas, schema semantics, corpus counts, or identifier conventions, as distinct from task-answer values.
  \item \textbf{Located values.} Specific data values, computed results, verdicts, or recommendations recorded with their location or provenance.
  \item \textbf{Test anticipation.} Explicit modeling of likely test asks, including predicted prompts and pre-answered probes. Generic descriptions of the task genre do not count.
  \item \textbf{Trap and contradiction documentation.} Seeded errors, dirty data, conflicting versions or figures, version-supersession doctrine, or adjudication guidance.
  \item \textbf{Verification doctrine.} Verify-live rules, anti-fabrication warnings, or re-derive-before-quoting requirements.
  \item \textbf{Format and grading guidance.} Output-format laws, tolerance or precision semantics, or grading mechanics translated into solver behavior. In-corpus scoring worksheets do not count.
  \item \textbf{Negative maps and budget economics.} What is absent, empty, broken, or not worth reading, and measured cost models of operations.
  \item \textbf{Environment corrections.} Overriding the environment's own documentation or naive expectations, including data repair and tool-capability matrices.
  \item \textbf{Identity and social decoding.} Persona orientation, role or alias maps, or name-collision warnings.
  \item \textbf{Honesty disclosures.} Staleness, coverage, or truncation caveats about the artifact itself, declared unknowns, or provenance labels.
\end{itemize}

\paragraph{Artifact labeling.} An LLM coder (\texttt{claude-fable-5}) labels each of the 216 artifact sets against the fixed category definitions. Corpus dumps inside \corpusToSkill{} outputs are excluded, and compiled skill trees are skimmed as machine-generated output. Each positive label requires a verbatim evidence quote. All 2{,}199 positive labels passed a mechanical check that the quote appears in the cited file. A second, independent LLM coder re-labeled a stratified sample of 24 artifact sets, covering 288 category decisions. Agreement was 95.5\% (275 of 288). Two disagreements affecting small-sample cells were adjudicated against the codebook. We treat the resulting counts as descriptive.

\begin{table}[h]
\centering
\scriptsize
\resizebox{\textwidth}{!}{%
\begin{tabular}{@{}l*{12}{c}@{}}
\toprule
& \multicolumn{2}{c}{BCP-G} & \multicolumn{2}{c}{OfficeQA} & \multicolumn{2}{c}{Harvey LAB} & \multicolumn{2}{c}{DABStep} & \multicolumn{2}{c}{Apex Agents} & \multicolumn{2}{c}{AppWorld} \\
\cmidrule(lr){2-3}\cmidrule(lr){4-5}\cmidrule(lr){6-7}\cmidrule(lr){8-9}\cmidrule(lr){10-11}\cmidrule(lr){12-13}
Artifact category & w/o & w/ & w/o & w/ & w/o & w/ & w/o & w/ & w/o & w/ & w/o & w/ \\
\midrule
Navigation architecture             & 3/3 & 3/3 & 3/3 & 3/3 & 3/3 & 3/3 & 3/3 & 3/3 & 93/93 & 93/93 & 3/3 & 3/3 \\
Method recipes                      & 3/3 & 3/3 & 3/3 & 3/3 & 3/3 & 3/3 & 3/3 & 3/3 & 93/93 & 92/93 & 3/3 & 3/3 \\
Environment constants               & 3/3 & 3/3 & 3/3 & 3/3 & 3/3 & 3/3 & 3/3 & 3/3 & 93/93 & 93/93 & 3/3 & 3/3 \\
Located values                      & 0/3 & 3/3 & 0/3 & 2/3 & 3/3 & 3/3 & 3/3 & 2/3 & 93/93 & 93/93 & 0/3 & 0/3 \\
Test anticipation                   & 1/3 & 1/3 & 2/3 & 3/3 & 2/3 & 1/3 & 3/3 & 3/3 & 60/93 & 61/93 & 3/3 & 1/3 \\
Trap \& contradiction documentation & 3/3 & 3/3 & 3/3 & 3/3 & 3/3 & 3/3 & 3/3 & 3/3 & 92/93 & 83/93 & 0/3 & 1/3 \\
Verification doctrine               & 1/3 & 3/3 & 3/3 & 3/3 & 3/3 & 3/3 & 2/3 & 3/3 & 34/93 & 93/93 & 1/3 & 3/3 \\
Format \& grading guidance          & 2/3 & 3/3 & 3/3 & 3/3 & 0/3 & 3/3 & 3/3 & 3/3 &  9/93 & 58/93 & 3/3 & 3/3 \\
Negative maps \& budget economics   & 3/3 & 3/3 & 3/3 & 3/3 & 3/3 & 3/3 & 3/3 & 3/3 & 93/93 & 75/93 & 3/3 & 2/3 \\
Environment corrections             & 3/3 & 3/3 & 3/3 & 3/3 & 3/3 & 3/3 & 3/3 & 3/3 & 92/93 & 74/93 & 3/3 & 3/3 \\
Identity \& social decoding         & 0/3 & 0/3 & 0/3 & 0/3 & 3/3 & 3/3 & 0/3 & 0/3 & 92/93 & 93/93 & 2/3 & 3/3 \\
Honesty disclosures                 & 1/3 & 3/3 & 3/3 & 3/3 & 3/3 & 3/3 & 1/3 & 3/3 & 59/93 & 91/93 & 2/3 & 3/3 \\
\bottomrule
\end{tabular}}
\caption{Artifact content produced by the \metaAgent{} variants. Each cell reports artifact sets exhibiting the category out of all artifact sets for that variant. \metaNoArchive{} and \metaWithArchive{} denote the variants without and with the workflow archive, respectively. Iterations are pooled, and Apex Agents is additionally pooled across worlds.}
\label{tab:artifact-taxonomy}
\end{table}

\paragraph{Artifact findings.} Navigation structures, method recipes, and environment constants appear in nearly every artifact set from both variants. Other content reflects the environment: identity and social decoding is concentrated in Harvey LAB and Apex Agents, while located values are absent from AppWorld, whose state is randomized across tasks. Archive-assisted artifacts also contain more operating guidance on Apex Agents. Verification guidance increases from 34 of 93 unaided artifacts to all 93 archive-assisted artifacts, format and grading guidance from 9 to 58, and honesty disclosures from 59 to 91. These descriptive patterns do not establish which artifact properties caused changes in downstream reward.

\subsection{Qualitative Artifact Examples}
\label{app:artifact-examples}

These excerpts illustrate two forms of environment-specific information captured during study: corrections to domain rules and consolidation of context scattered across the environment. Both were produced by \metaWithArchive{} in iteration one. Typography is normalized, and bracketed ellipses mark omitted text.

\paragraph{DABStep: correcting documentation and recovering computation rules.}
The artifact corrects the supplied documentation and records schema and fee-computation rules that cannot be inferred safely from field names alone. The excerpt comes from the \texttt{payment-data} environment group at \path{/meta_archive/artifacts/cheatsheet/study/README.md}.
\begin{quote}
\footnotesize\ttfamily\raggedright
payments-readme.md lists card\_scheme as [MasterCard, Visa, Amex, Other]. WRONG. Real schemes = NexPay, GlobalCard, SwiftCharge, TransactPlus.\\[2pt]
[\ldots] fee = fixed\_amount + rate * transaction\_value / 10000.\\[2pt]
A rule applies to a txn only if EVERY field matches. Rule field null/empty-list = wildcard (matches all).\\[2pt]
monthly\_volume and monthly\_fraud\_level are not stored. They must be COMPUTED per merchant per calendar month from payments, then matched to the rule's range string.
\end{quote}

\paragraph{Apex Agents: reconstructing identities and decision rationale.}
The artifact links inconsistent names across communication channels and preserves the rationale behind a project decision. The excerpt comes from \texttt{investment-banking-world-221} at \path{/meta_archive/artifacts/cheatsheet/study/02_people_and_narrative.md}.
\begin{quote}
\footnotesize\ttfamily\raggedright
Ola / ``Tarek'' Amethyst: Deal structuring, M\&A \& pro forma analysis (chat msg 71 addresses ``Tarek'' for M\&A slides = same person as ``Ola'').\\[2pt]
Huajian / Luca Amethyst: Pitch deck. ``Luca'' does the slide work in chat.\\[2pt]
Names drift: Ola $\approx$ Tarek, and Huajian $\approx$ Luca. The email From names and chat names don't perfectly align. Treat by ROLE, not name.\\[2pt]
[\ldots] Final target = TPVG [\ldots] despite Luca flagging TPVG is venture/growth-focused, outside BBDC's senior-secured profile. Justification: diversification + tech/venture exposure.\\[2pt]
No DCF for BBDC. Valuation = Trading Comps only. Merger model = accretion/dilution.
\end{quote}

The examples show how study artifacts can make both domain rules and dispersed contextual information directly available to the solver.

\section{Test-Time Interpretability}
\label{app:test-interpretability}

\subsection{Observed Artifact Use}
\label{app:artifact-use-lift}

Determining whether an artifact influenced a rollout is inherently difficult. Artifacts may be accessed through explicit file reads or supplied through prompts, and a solver may act on injected guidance without visibly retrieving it. We therefore measure \emph{observed artifact use}, not causal influence.

\paragraph{Trace labeling.}
We use \texttt{gpt-5.6-luna} to classify task-time traces. A trace counts as observed use when the solver deliberately retrieves artifact content or clearly acts on distinctive guidance from it. Mere exposure, directory listings, failed retrievals, and generic behavior do not count. The judge receives the trajectory and mechanical annotations of artifact-related actions, but not its reward, corresponding \noStudy{} traces, or verifier output. Figures~\ref{fig:artifact-use-rubric} and~\ref{fig:artifact-use-rubric-gates} provide the complete rubric.

Because artifact delivery differs across methods, ``no observed use'' does not establish that an artifact had no influence, and use rates should not be compared across methods.

\paragraph{Within-task comparison.}
For each task, method, and artifact iteration represented by both observed-use and no-observed-use trials, we compute the difference in mean reward between the two groups and average these differences across tasks. We omit cells with fewer than 10 eligible task--iterations. This controls for differences between tasks but remains observational because artifact use may follow events within a rollout.

\begin{table}[h]
\centering
\scriptsize
\setlength{\tabcolsep}{3pt}
\begin{tabular*}{\textwidth}{@{\extracolsep{\fill}}lcccccc@{}}
\toprule
Method & BCP-G & OfficeQA & Harvey LAB & DABStep & Apex Agents & AppWorld \\
\midrule
\preping{} & $-$0.054 & +0.000 & +0.005 & +0.071 & $-$0.005 & $-$0.009 \\
\corpusToSkill{} & -- & -- & -- & $-$0.032 & +0.020 & $-$0.006 \\
\metaNoArchive{} & +0.015 & -- & +0.029 & +0.112 & +0.038 & -- \\
\metaWithArchive{} & $-$0.044 & $-$0.005 & -- & +0.080 & $-$0.010 & $-$0.042 \\
\bottomrule
\end{tabular*}
\caption{Within-task reward difference between observed-use and no-observed-use trials. Dashes indicate fewer than 10 contributing task--iterations. Full coverage counts will be included in the released analysis tables.}
\label{tab:artifact-use-within-task}
\end{table}

Observed use is not uniformly associated with higher reward. Among the within-task contrasts, the clearest positive associations occur on DABStep for \preping{} and both \metaAgent{} variants. The BCP-G contrasts for \preping{} and \metaWithArchive{} are negative. These patterns are descriptive and do not isolate the causal effect of artifact use.

\input{figures/fig_artifact_use_rubric.tex}
\input{figures/fig_artifact_use_rubric_gates.tex}

\subsection{Lift by Task Stratum}
\label{app:strata-lift}

We next ask which kinds of tasks benefit from studying. We use benchmark-provided difficulty labels for OfficeQA~\citep{officeqa} and DABStep~\citep{dabstep}, difficulty and application-count labels for AppWorld~\citep{appworld}, domain labels for Apex Agents~\citep{apexagents}, and topic and gold-document counts for BCP-G~\citep{browsecompplus}. Exact metadata sources and mappings will accompany the released analysis.

For each task stratum $S$, we report
\begin{equation}
    \frac{1}{|S|}\sum_{i \in S}\left(r_i - \bar{r}^{\mathrm{NS}}_{t(i)}\right),
    \label{eq:stratum-lift}
\end{equation}
where $\bar{r}^{\mathrm{NS}}_{t(i)}$ is the mean reward from a fixed three-run subset of \noStudy{} evaluations for task $t(i)$. We use this subset for computational reasons. The primary results use all ten \noStudy{} runs. Artifact iterations are pooled. This analysis is exploratory, and the comparisons are not adjusted for the number of strata examined.

\begin{table}[h]
\centering
\small
\setlength{\tabcolsep}{2.7pt}
\begin{tabular*}{\textwidth}{@{\extracolsep{\fill}}lccccccc@{}}
\toprule
& \multicolumn{2}{c}{OfficeQA} & \multicolumn{2}{c}{DABStep} & \multicolumn{3}{c}{AppWorld} \\
\cmidrule(lr){2-3}\cmidrule(lr){4-5}\cmidrule(lr){6-8}
Method & Easy & Hard & Easy & Hard & Difficulty 1 & Difficulty 2 & Difficulty 3 \\
\midrule
\noStudy{} reward & 0.735 & 0.459 & 0.792 & 0.293 & 0.731 & 0.465 & 0.492 \\
\addlinespace[2pt]
\preping{} & +0.031 & +0.017 & $-$0.012 & +0.111 & $-$0.002 & +0.183 & +0.074 \\
\corpusToSkill{} & $-$0.050 & $-$0.050 & +0.000 & $-$0.058 & $-$0.023 & +0.049 & +0.085 \\
\metaNoArchive{} & +0.046 & +0.049 & +0.009 & +0.030 & +0.101 & +0.134 & +0.134 \\
\metaWithArchive{} & +0.038 & +0.018 & +0.015 & +0.185 & +0.074 & +0.139 & +0.092 \\
\bottomrule
\end{tabular*}
\caption{Task-matched reward lift over \noStudy{} by dataset difficulty label. The first row gives the \noStudy{} reward level of each stratum. Full counts and intervals are provided in the released analysis. AppWorld uses its three-level difficulty scale.}
\label{tab:strata-difficulty}
\end{table}

\paragraph{Difficulty.}
Difficulty does not produce a universal pattern. On DABStep, the positive estimates for \preping{} and both \metaAgent{} variants are substantially larger on hard tasks than on easy ones. \metaWithArchive{} improves reward by 0.185 on hard tasks but only 0.015 on easy tasks, where \noStudy{} already scores 0.792. OfficeQA shows little separation between difficulty levels, and AppWorld's gains are positive but non-monotonic for the \metaAgent{} variants.

\begin{table}[h]
\centering
\small
\begin{tabular*}{\textwidth}{@{\extracolsep{\fill}}lccc@{}}
\toprule
Method & Investment Banking & Law & Management Consulting \\
\midrule
\noStudy{} reward & 0.082 & 0.247 & 0.203 \\
\addlinespace[2pt]
\preping{} & +0.009 & +0.089 & +0.035 \\
\corpusToSkill{} & +0.001 & +0.122 & $-$0.017 \\
\metaNoArchive{} & $-$0.004 & +0.113 & +0.003 \\
\metaWithArchive{} & +0.002 & +0.139 & 0.000 \\
\bottomrule
\end{tabular*}
\caption{Task-matched reward lift over \noStudy{} on Apex Agents by task domain. The first row gives the \noStudy{} reward level of each domain. Full counts and intervals are provided in the released analysis.}
\label{tab:strata-apex-domain}
\end{table}

\paragraph{Apex Agents gains concentrate in Law.}
All four methods have their largest domain-level lift on Law (+0.089 to +0.139). Estimates are near zero in Investment Banking ($-$0.004 to +0.009) and mixed in Management Consulting ($-$0.017 to +0.035).

\paragraph{BCP-G retrieval strata.}
\corpusToSkill{} has positive lift across all ten topic labels (Table~\ref{tab:strata-bcp-topic}), indicating that its overall gain is not driven by a single topic. Its lift also rises from +0.224 with one gold document to +0.293 with four or more (Table~\ref{tab:strata-bcp-documents}). This pattern is consistent with corpus compilation being more useful when tasks require locating more supporting documents, although the analysis does not isolate retrieval burden from other task differences.

\paragraph{AppWorld application count.}
Every studying method has its largest estimated lift on tasks involving two applications (Table~\ref{tab:strata-appworld-scale}), with reward gains from +0.108 to +0.141.

Together, these results do not reveal a universal notion of task difficulty. Studying helps when the information captured during study matches the structure required by downstream tasks.

\input{sections/appendix_strata_tables.tex}
\clearpage

\subsection{Helpful and Harmful Artifact Use}
\label{app:test-trace-examples}

Two matched examples illustrate how artifacts can help or hinder the solver. Each compares a \metaWithArchive{} rollout with a \noStudy{} rollout on the same task and downstream model. The examples are not paired by random seed and should not be interpreted causally.

\paragraph{DABStep: recovering the correct fraud boundary.}
For task \texttt{1701}, the artifact records the dataset's fraud-level buckets:
\begin{quote}
\small\ttfamily
monthly\_fraud\_level buckets: <7.2\%, 7.2\%--7.7\%, 7.7\%--8.3\%, >8.3\%.\\
= fraud volume / total volume that month.
\end{quote}
The archive-assisted solver reads this guidance and applies the correct upper bucket:
\begin{quote}
\small\ttfamily
Read /meta\_archive/artifacts/cheatsheet/study/fee\_matching.md\\
April 2023 fraud rate: 9.92\%\\
Fraud level bucket: >8.3\%
\end{quote}
The matched \noStudy{} solver computes the same fraud rate but invents a different boundary:
\begin{quote}
\small\ttfamily
elif fraud\_rate <= 9: fraud\_bracket = `8\%--9\%'\\
else: fraud\_bracket = `>9\%'\\
monthly\_fraud\_level = `>9\%'
\end{quote}
The incorrect bucket propagates into the final Fee-ID set: \noStudy{} fails all 10 evaluations on this task, while the selected artifact succeeds in all three of its evaluations.

\paragraph{Apex Agents: an artifact creates a search tunnel.}
For task \texttt{word-129-pj-01}, the artifact directs the solver toward general ServiceNow pricing documents:
\begin{quote}
\small\ttfamily
These documents collectively cover the competitive pricing, packaging, and market positioning landscape for workflow automation platforms \ldots{} [including] ServiceNow.\\
Competitions\_Price\_Benchmarking.xlsx, \ldots{} ServiceNow ITOM HRSD SecOps modules.
\end{quote}

The solver follows this route and substitutes unsupported assumptions for the task inputs:
\begin{quote}
\small\ttfamily
The skills that seem most relevant are:\\
skill-02-workflow-automation-pricing \ldots{}\\
Assumptions for calculation: \ldots{} Total ServiceNow customers: 1,000 \ldots{} Average users per customer: 150 \ldots{} Module adoption varies by customer segment.\\
ServiceNow Total Annual Revenue Forecast: \$146.2m
\end{quote}

The artifact does not index the two root-level images containing the required subscriber counts and price multipliers. The matched \noStudy{} solver discovers those files through a directory listing and obtains the correct total of \$1{,}438.2 million. \noStudy{} succeeds in 6 of 10 evaluations on this task, whereas \metaWithArchive{} fails in all nine evaluations across the three artifact iterations. This example shows how a plausible but incomplete artifact can narrow the solver's search and hide relevant evidence.

%% file: figures/fig_study_files_vs_score.tex
\begin{figure}[h!]
\centering
\includegraphics[width=\textwidth]{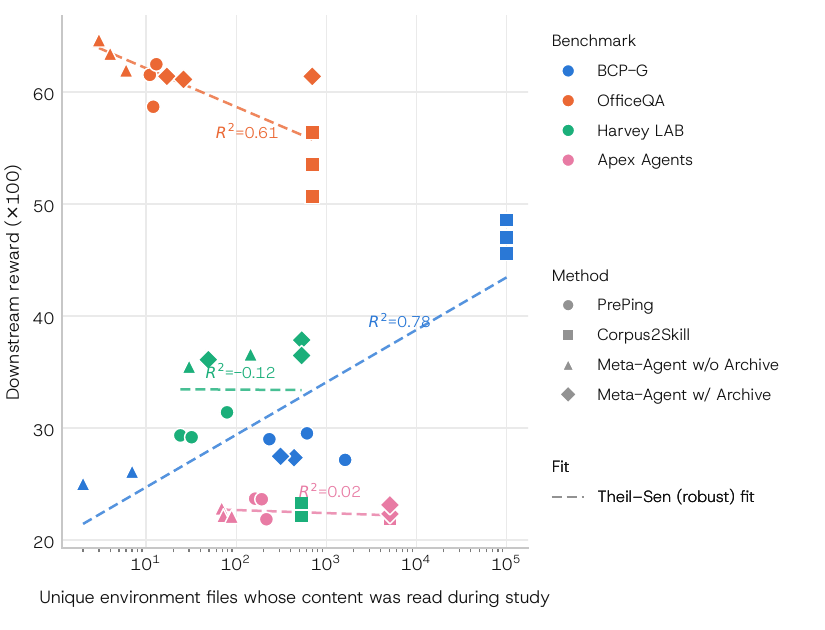}
\caption{\textbf{File access and downstream reward.} Each point is one benchmark--method--artifact-iteration combination. The horizontal axis counts unique environment files accessed during the analyzed study attempt. The vertical axis is that artifact's task-macro Avg@3 reward. Dashed lines are descriptive per-benchmark Theil--Sen fits of reward on $\log_{10}$(files), pooled across methods and iterations. The annotated $R^2$ is computed against the robust line. DABStep is omitted because all methods access all seven files. AppWorld has no real file surface.}
\label{fig:study-files-vs-score}
\end{figure}

%% file: figures/fig_study_tools_vs_score.tex
\begin{figure}[t]
\centering
\includegraphics[width=\textwidth]{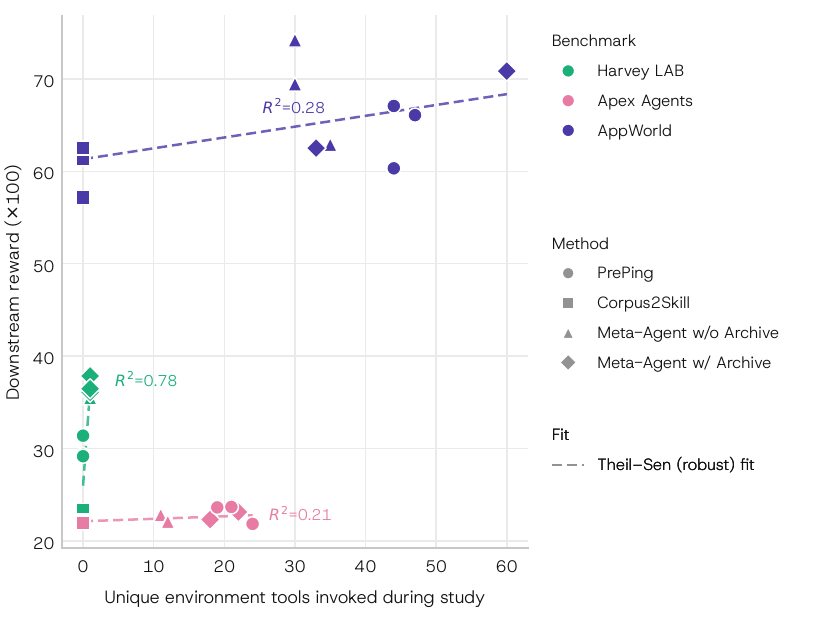}
\caption{\textbf{Tool use and downstream reward.} Each point is one benchmark--method--artifact-iteration combination. The horizontal axis counts distinct benchmark-specific tools invoked during the analyzed study attempt. The vertical axis is that artifact's task-macro Avg@3 reward. Dashed lines are descriptive per-benchmark Theil--Sen fits pooled across methods and iterations, with $R^2$ computed against the robust line. Harvey LAB has one benchmark-specific tool, so its fit represents a method-confounded binary split rather than a continuous breadth relationship.}
\label{fig:study-tools-vs-score}
\end{figure}

%% file: figures/fig_artifact_use_rubric.tex
\begin{figure}[p]
\centering
\begin{minipage}{\textwidth}
\scriptsize
\begin{verbatim}
You are auditing whether a test-time solver agent USED a study artifact that was placed in its
environment. You will see the solver's trace and mechanical annotations. You are NOT judging
whether the artifact helped -- only whether the agent used it.

Output exactly one JSON object:
{"used": true|false,
 "channel": "<how the artifact was engaged: e.g. 'read /cheatsheet files', 'compiled
             get_document', 'playbook citation', or 'none'>",
 "evidence": ["2-5 bullets citing step ids and short quotes"],
 "confidence": 0.0-1.0}

USED (true) means at least one of:
- The agent, by its own tool call, brought artifact CONTENT into its context: a
  Read/cat/grep/head of a file under an artifact mount, a Skill load, or a SUCCESSFUL
  get_document call -- and the retrieved content is more than a bare directory listing.
- The agent's OWN authored text (message, reasoning, or tool arguments -- never user-role text
  or tool observations) engages injected artifact content: it cites a playbook ID (e.g.
  [strategies-NNN]/[pitfalls-NNN]), explicitly invokes the artifact ("According to the
  playbook, I should ..." / "Based on the playbook, I expect ...") and then acts on the stated
  plan -- this counts even when the plan's individual tactics are generic, because the agent is
  visibly steering by the artifact -- or reproduces a distinctive artifact-only fact, rule, or
  ORDERED multi-step recipe at a point where it shapes what the agent does next. Compare the
  agent's behavior against the injected playbook text you can see in the first user message: a
  specific fact or rule stated in the playbook and applied by the agent counts even without a
  citation. A single generic tactic (grep, read-the-docs) that any competent agent might use
  does not.

NOT used (false):
- Exposure alone: injected playbook/prefix text sitting in the prompt, or artifact text echoed
  back in observations, with no agent-authored engagement.
- A coerced or trivial touch: an `ls` of the mount, a mandated Skill call whose output the
  agent never draws on, or get_document calls that ALL errored.
- Generic diligence that matches artifact advice only thematically, with no citation and no
  artifact-only fact (model-default behavior).

Hard rules:
1. Only agent-authored text counts for citations/provenance. Artifact text arrives in user-role
   messages and large read observations -- matches there are exposure, not use.
2. Echo traps: /logs/agent dumps, tool-result re-reads, and embedded subagent transcripts do
   not count as the agent engaging the artifact.
3. Reading an artifact file and then visibly ignoring it still counts as used (content entered
   context by the agent's own action). Merely listing filenames does not.
4. When genuinely uncertain, set used=false with low confidence rather than guessing true.
\end{verbatim}
\end{minipage}
\caption{Artifact-use judge rubric (part one of two): core criteria. Each prompt also includes the
relevant method gate and benchmark note from Figure~\ref{fig:artifact-use-rubric-gates},
mechanical annotations, and the rendered trace. Method gates take precedence over the core
criteria.}
\label{fig:artifact-use-rubric}
\end{figure}

%% file: figures/fig_artifact_use_rubric_gates.tex
\begin{figure}[p]
\centering
\begin{minipage}{\textwidth}
\scriptsize
\begin{verbatim}
METHOD c2s: the corpus is compiled into navigable skills (Skill tool, INDEX/SKILL.md files, MCP
corpus-documents get_document, mounted under /opt/corpus2skill or similar). On most benchmarks
a system prompt MANDATES >=2 Skill calls and >=1 get_document. A mandated call still counts as
used ONLY if artifact content actually entered context (successful get_document, or index/skill
text visibly consulted beyond the coerced minimum). All-error get_document with no
skill-content engagement is not use.

METHOD cheatsheet: small markdown files mounted read-only (at /cheatsheet or /context). The
task preamble tells the agent it may inspect them. Used = the agent Read/cat/grep'd cheatsheet
file CONTENT (tool args show the path). A bare `ls` of the mount, or the mount merely being
named in prompts, is not use.

METHOD preping: a playbook is pasted into the FIRST USER MESSAGE (PLAYBOOK_BEGIN/END, bullets
tagged [strategies-NNN]/[pitfalls-NNN]/[apis-NNN]/[code_snippets-NNN]). Exposure is 100% by
construction. Only agent-authored engagement counts. Used = a playbook-ID citation in agent
text, OR agent text/code reproducing a distinctive playbook-only fact or recipe (a specific
file/table/endpoint the playbook names, a code snippet's distinctive shape) at a point that
shapes its next action. Generic strategies the model would use anyway are not use.

METHOD meta-agent-v1: a study bundle is delivered two ways at once -- (a) an orientation prefix
injected into the SYSTEM PROMPT (not visible in the trace, and it may embed a playbook with
[strategies-NNN]-style tags and names artifact file locations), and (b) artifact files mounted
under /meta_archive/... (e.g. /meta_archive/artifacts/cheatsheet/study/,
/meta_archive/artifacts/preping/playbook/). Used = the agent reads /meta_archive file content,
OR its own text cites a playbook ID or reproduces a distinctive bundle-only fact or recipe
(e.g. a corpus-era/table-format rule, a named source file, a prescribed search recipe) at a
decision point. Because the prefix is invisible here, weigh the mechanical annotations and
distinctive-content evidence carefully. Generic competence is not use.

Benchmark appworld: the agent drives simulated apps via mcp__appworld__* tools. c2s here has no
system mandate. Any skill/get_document access was a free choice. Cheatsheet files live under
/cheatsheet (e.g. 00_START_HERE.md, per-app API notes).

Benchmark apex: finance/law/consulting tasks over /filesystem and /.apps_data. c2s get_document
is often broken ('not connected'/'No such tool'). All-error access is not use. The cheatsheet
is a per-world markdown pack. The preping playbook often names exact workbook paths, tab
indices, cell ranges, or figures. An agent going straight to a playbook-named path/tab or
quoting playbook-supplied figures is using the playbook even without a citation.

Benchmark bcp-grep (BrowseComp over plaintext /workspace/corpus with native Grep/Glob/Read).
The cheatsheet README carries search mechanics (rg flags, triage recipes). c2s replaces search
with skills+get_document, so c2s corpus access is itself artifact use when content is
retrieved. The preping playbook prescribes an ORDERED candidate-list intersection recipe
(corpus-wide grep -rl for one clue, intersect with a second clue via xargs grep -l, then read
the intersection) -- an agent executing that ordered recipe is using the playbook even without
a citation. A lone grep is not.

Benchmark dabstep: data analysis over /app/data (payments.csv + docs). c2s compiles the same
files the agent can read directly. Only reads of the COMPILED copies or successful get_document
count as artifact use, reads of /app/data originals do not. Known distinctive playbook-only
rule: treating an EMPTY LIST in fees.json fields (account_type, aci) as a wildcard applying to
all values -- the corpus manual only covers null. An agent stating or applying
empty-list-as-wildcard is using the playbook.

Benchmark harvey_lab: legal research over a 9,288-doc DMS at /dms (labread CLI for binary
docs). Cheatsheet = 00-START-HERE.md + matter-index.md under the artifact mount. Reads of /dms
originals are corpus access, not artifact use.

Benchmark officeqa: numeric answers from 697 Treasury Bulletin text files at /app/corpus.
Artifact mounts hold corpus maps and table-parsing recipes. Reads of /app/corpus originals are
corpus access, not artifact use.
\end{verbatim}
\end{minipage}
\caption{Artifact-use judge rubric (part two of two): method gates and benchmark notes. Each prompt
includes one of each. The \corpusToSkill{} gate excludes mandated calls that do not bring
artifact content into context.}
\label{fig:artifact-use-rubric-gates}
\end{figure}

%% file: sections/appendix_strata_tables.tex
\begin{table}[t]
\centering
\small
\begin{tabular*}{\textwidth}{@{\extracolsep{\fill}}lccc@{}}
\toprule
Method & 1 app & 2 apps & 3 apps \\
\midrule
\noStudy{} reward & 0.684 & 0.444 & 0.556 \\
\addlinespace[2pt]
\preping{} & +0.036 & +0.141 & +0.000 \\
\corpusToSkill{} & $-$0.026 & +0.108 & +0.015 \\
\metaNoArchive{} & +0.131 & +0.133 & +0.030 \\
\metaWithArchive{} & +0.077 & +0.132 & +0.052 \\
\bottomrule
\end{tabular*}
\caption{Task-matched reward lift over \noStudy{} on AppWorld by the number of applications required. The first row gives the \noStudy{} reward level of each stratum. Full counts and intervals are provided in the released analysis.}
\label{tab:strata-appworld-scale}
\end{table}

\begin{table}[t]
\centering
\small
\begin{tabular*}{\textwidth}{@{\extracolsep{\fill}}lccc@{}}
\toprule
Method & 1 gold document & 2--3 gold documents & 4+ gold documents \\
\midrule
\noStudy{} reward & 0.226 & 0.208 & 0.203 \\
\addlinespace[2pt]
\preping{} & +0.086 & +0.060 & +0.078 \\
\corpusToSkill{} & +0.224 & +0.258 & +0.293 \\
\metaNoArchive{} & +0.048 & +0.027 & +0.030 \\
\metaWithArchive{} & +0.092 & +0.068 & +0.063 \\
\bottomrule
\end{tabular*}
\caption{Task-matched reward lift over \noStudy{} on BCP-G by the number of gold documents. The first row gives the \noStudy{} reward level of each stratum. Full counts and intervals are provided in the released analysis.}
\label{tab:strata-bcp-documents}
\end{table}

\begin{table}[t]
\centering
\scriptsize
\setlength{\tabcolsep}{2.5pt}
\resizebox{\textwidth}{!}{%
\begin{tabular}{@{}lcccccccccc@{}}
\toprule
 & \multicolumn{10}{c}{Topic} \\
\cmidrule(lr){2-11}
Method & Art & Geography & History & Music & Politics & Science \& tech. & Sports & TV \& movies & Video games & Other \\
\midrule
\noStudy{} reward & 0.246 & 0.161 & 0.187 & 0.173 & 0.250 & 0.365 & 0.165 & 0.135 & 0.153 & 0.248 \\
\addlinespace[2pt]
\preping{} & +0.063 & +0.099 & +0.075 & +0.093 & +0.085 & +0.062 & +0.072 & +0.051 & +0.069 & +0.104 \\
\corpusToSkill{} & +0.228 & +0.353 & +0.226 & +0.321 & +0.188 & +0.188 & +0.381 & +0.242 & +0.373 & +0.197 \\
\metaNoArchive{} & +0.037 & +0.059 & +0.027 & +0.029 & +0.045 & +0.037 & +0.036 & +0.026 & +0.029 & +0.043 \\
\metaWithArchive{} & +0.046 & +0.087 & +0.078 & +0.082 & +0.108 & +0.102 & +0.060 & +0.054 & +0.073 & +0.082 \\
\bottomrule
\end{tabular}}
\caption{Task-matched reward lift over \noStudy{} on BCP-G by BrowseComp topic label. The first row gives the \noStudy{} reward level of each topic. Full counts and intervals are provided in the released analysis.}
\label{tab:strata-bcp-topic}
\end{table}

%% file: references.bib
@misc{yao2026harnessbenchmeasuringharnesseffects,
      title={Harness-Bench: Measuring Harness Effects across Models in Realistic Agent Workflows}, 
      author={Yilun Yao and Xinyu Tan and Chao-Hsuan Liu and Yaoming Li and Zhengyang Wang and Wenhan Yu and Zhewen Tan and Yuxuan Tian and Guangxiang Zhao and Lin Sun and Xiangzheng Zhang and Tong Yang},
      year={2026},
      eprint={2605.27922},
      archivePrefix={arXiv},
      primaryClass={cs.AI},
      url={https://arxiv.org/abs/2605.27922}, 
}

@misc{anthropic_claude_code_docs,
  author       = {{Anthropic}},
  title        = {Claude Code Documentation},
  year         = {2026},
  howpublished = {\url{https://code.claude.com/docs/en/overview}},
  note         = {Accessed: 2026-09-01}
}

@misc{Harbor_Framework,
author = {{Harbor Framework Team}},
title = {{Harbor: A framework for evaluating and optimizing agents and models in container environments}},
year = {2026},
howpublished = {Zenodo},
doi = {10.5281/zenodo.20953922},
url = {https://doi.org/10.5281/zenodo.20953922}
}

@inproceedings{
ursekar2026vero,
title={Ve{RO}: A Harness for Agents to Optimize Agents},
author={Varun Ursekar and Apaar Shanker and Veronica Chatrath and Yuan Xue and Samuel Marc Denton},
booktitle={Forty-third International Conference on Machine Learning},
year={2026},
url={https://openreview.net/forum?id=zQzmwG2Nue}
}

@inproceedings{lee2026metaharness,
  title={Meta-Harness: End-to-End Optimization of Model Harnesses},
  author={Lee, Yoonho and Nair, Roshen and Zhang, Qizheng and Lee, Kangwook and Khattab, Omar and Finn, Chelsea},
  booktitle={Conference on Language Modeling (COLM)},
  year={2026}
}

@article{weng2026harness,
  title = {Harness Engineering for Self-Improvement},
  author = {Weng, Lilian},
  journal = {lilianweng.github.io},
  year = {2026},
  month = {July},
  url = "https://lilianweng.github.io/posts/2026-07-04-harness/"
}

@misc{liu2025spiceselfplaycorpusenvironments,
      title={SPICE: Self-Play In Corpus Environments Improves Reasoning}, 
      author={Bo Liu and Chuanyang Jin and Seungone Kim and Weizhe Yuan and Wenting Zhao and Ilia Kulikov and Xian Li and Sainbayar Sukhbaatar and Jack Lanchantin and Jason Weston},
      year={2025},
      eprint={2510.24684},
      archivePrefix={arXiv},
      primaryClass={cs.CL},
      url={https://arxiv.org/abs/2510.24684}, 
}

@inproceedings{sweagent,
  title     = {{SWE-agent}: Agent-Computer Interfaces Enable Automated Software Engineering},
  author    = {Yang, John and Jimenez, Carlos E. and Wettig, Alexander and Lieret, Kilian and Yao, Shunyu and Narasimhan, Karthik and Press, Ofir},
  booktitle = {Advances in Neural Information Processing Systems (NeurIPS)},
  year      = {2024},
  doi       = {10.48550/arXiv.2405.15793},
  note      = {arXiv:2405.15793}
}

@article{browsergymecosystem,
  title   = {The {BrowserGym} Ecosystem for Web Agent Research},
  author  = {Le Sellier De Chezelles, Thibault and Gasse, Maxime and Drouin, Alexandre and Caccia, Massimo and Boisvert, L{\'e}o and Thakkar, Megh and Marty, Tom and Assouel, Rim and Shayegan, Sahar Omidi and Jang, Lawrence Keunho and L{\`u}, Xing Han and Yoran, Ori and Kong, Dehan and Xu, Frank F. and Reddy, Siva and Cappart, Quentin and Neubig, Graham and Salakhutdinov, Ruslan and Chapados, Nicolas and Lacoste, Alexandre},
  journal = {arXiv preprint arXiv:2412.05467},
  year    = {2024},
  doi     = {10.48550/arXiv.2412.05467}
}

@article{preping,
  title   = {{PREPING}: Building Agent Memory without Tasks},
  author  = {Choi, Yumin and Park, Sangwoo and Kang, Minki and Baek, Jinheon and Hwang, Sung Ju},
  journal = {arXiv preprint arXiv:2605.13880},
  year    = {2026},
  doi     = {10.48550/arXiv.2605.13880}
}

@article{corpus2skill,
  title   = {{Corpus2Skill}: Distilling Enterprise Knowledge into Navigable Agent Skills for {QA} and {RAG}},
  author  = {Sun, Yiqun and Wei, Pengfei and Hsieh, Lawrence B.},
  journal = {arXiv preprint arXiv:2604.14572},
  year    = {2026},
  doi     = {10.48550/arXiv.2604.14572},
  note    = {Accepted to EMNLP 2026 (Findings). Titled ``Don't Retrieve, Navigate'' in v1}
}

@article{voyager,
  title   = {Voyager: An Open-Ended Embodied Agent with Large Language Models},
  author  = {Wang, Guanzhi and Xie, Yuqi and Jiang, Yunfan and Mandlekar, Ajay and Xiao, Chaowei and Zhu, Yuke and Fan, Linxi and Anandkumar, Anima},
  journal = {arXiv preprint arXiv:2305.16291},
  year    = {2023},
  doi     = {10.48550/arXiv.2305.16291}
}

@article{awm,
  title   = {Agent Workflow Memory},
  author  = {Wang, Zora Zhiruo and Mao, Jiayuan and Fried, Daniel and Neubig, Graham},
  journal = {arXiv preprint arXiv:2409.07429},
  year    = {2024},
  doi     = {10.48550/arXiv.2409.07429},
  note    = {Published at ICML 2025 (PMLR v267, pp.\ 63897--63911)}
}

@article{memp,
  title   = {{Memp}: Exploring Agent Procedural Memory},
  author  = {Fang, Runnan and Liang, Yuan and Wang, Xiaobin and Wu, Jialong and Qiao, Shuofei and Xie, Pengjun and Huang, Fei and Chen, Huajun and Zhang, Ningyu},
  journal = {arXiv preprint arXiv:2508.06433},
  year    = {2025},
  doi     = {10.48550/arXiv.2508.06433}
}

@inproceedings{reflexion,
  title     = {Reflexion: Language Agents with Verbal Reinforcement Learning},
  author    = {Shinn, Noah and Cassano, Federico and Gopinath, Ashwin and Narasimhan, Karthik and Yao, Shunyu},
  booktitle = {Advances in Neural Information Processing Systems (NeurIPS)},
  volume    = {36},
  pages     = {8634--8652},
  publisher = {Curran Associates, Inc.},
  year      = {2023},
  doi       = {10.52202/075280-0377}
}

@article{expel,
  title     = {{ExpeL}: {LLM} Agents Are Experiential Learners},
  author    = {Zhao, Andrew and Huang, Daniel and Xu, Quentin and Lin, Matthieu and Liu, Yong-Jin and Huang, Gao},
  journal   = {Proceedings of the AAAI Conference on Artificial Intelligence},
  volume    = {38},
  number    = {17},
  pages     = {19632--19642},
  year      = {2024},
  doi       = {10.1609/aaai.v38i17.29936}
}

@article{memgpt,
  title   = {{MemGPT}: Towards {LLMs} as Operating Systems},
  author  = {Packer, Charles and Wooders, Sarah and Lin, Kevin and Fang, Vivian and Patil, Shishir G. and Stoica, Ion and Gonzalez, Joseph E.},
  journal = {arXiv preprint arXiv:2310.08560},
  year    = {2023},
  doi     = {10.48550/arXiv.2310.08560}
}

@article{amem,
  title   = {{A-MEM}: Agentic Memory for {LLM} Agents},
  author  = {Xu, Wujiang and Liang, Zujie and Mei, Kai and Gao, Hang and Tan, Juntao and Zhang, Yongfeng},
  journal = {arXiv preprint arXiv:2502.12110},
  year    = {2025},
  doi     = {10.48550/arXiv.2502.12110},
  note    = {Accepted at NeurIPS 2025}
}

@article{mem0,
  title   = {{Mem0}: Building Production-Ready {AI} Agents with Scalable Long-Term Memory},
  author  = {Chhikara, Prateek and Khant, Dev and Aryan, Saket and Singh, Taranjeet and Yadav, Deshraj},
  journal = {arXiv preprint arXiv:2504.19413},
  year    = {2025},
  doi     = {10.48550/arXiv.2504.19413}
}

@article{dynamiccheatsheet,
  title   = {Dynamic Cheatsheet: Test-Time Learning with Adaptive Memory},
  author  = {Suzgun, Mirac and Y{\"u}ksekg{\"o}n{\"u}l, Mert and Bianchi, Federico and Jurafsky, Dan and Zou, James},
  journal = {arXiv preprint arXiv:2504.07952},
  year    = {2025},
  doi     = {10.48550/arXiv.2504.07952},
  note    = {Published at EACL 2026}
}

@article{ace,
  title   = {Agentic Context Engineering: Evolving Contexts for Self-Improving Language Models},
  author  = {Zhang, Qizheng and Hu, Changran and Upasani, Shubhangi and Ma, Boyuan and Hong, Fenglu and Kamanuru, Vamsidhar and Rainton, Jay and Wu, Chen and Ji, Mengmeng and Li, Hanchen and Thakker, Urmish and Zou, James and Olukotun, Kunle},
  journal = {arXiv preprint arXiv:2510.04618},
  year    = {2025},
  doi     = {10.48550/arXiv.2510.04618},
  note    = {Published at ICLR 2026}
}

@article{sleeptime,
  title   = {Sleep-Time Compute: Beyond Inference Scaling at Test-Time},
  author  = {Lin, K. and Snell, C. and Wang, Y. and Packer, C. and Wooders, S. and Stoica, I. and Gonzalez, J. E.},
  journal = {arXiv preprint arXiv:2504.13171},
  year    = {2025},
  doi     = {10.48550/arXiv.2504.13171}
}

@inproceedings{
machinestudying,
title={Machine Studying: A System-Level Reframing of Continual Adaptation from Declarative Corpora},
author={Jacob Xiaochen Li and Rick Battle and Omar Khattab},
booktitle={Continual Adaptation at Scale: Towards Sustainable AI Workshop},
year={2026},
url={https://openreview.net/forum?id=e7duOnFgL7}
}

@article{proact,
  title   = {Anticipate and Learn: Unleashing Idle-Time Compute in Proactive Agents},
  author  = {Hu, Haoyi and Lyu, Qirong and Kong, Xianghan and Liu, Weiwen and Lin, Jianghao and Guo, Zixuan and Xu, Yan and Wang, Yasheng and Zhang, Weinan and Yu, Yong},
  journal = {arXiv preprint arXiv:2605.25971},
  year    = {2026},
  doi     = {10.48550/arXiv.2605.25971}
}

@article{idlespec,
  title   = {{IdleSpec}: Exploiting Idle Time via Speculative Planning for {LLM} Agents},
  author  = {Choi, Daewon and Park, Kyunghyun and Song, Woomin and Dingliwal, Saket and Jayanthi, Sai Muralidhar and Shin, Jinwoo and Galstyan, Aram},
  journal = {arXiv preprint arXiv:2605.22154},
  year    = {2026},
  doi     = {10.48550/arXiv.2605.22154}
}

@article{autodreamer,
  title   = {{Auto-Dreamer}: Learning Offline Memory Consolidation for Language Agents},
  author  = {Ye, Chongrui and Liu, Yuxiang and Wang, Yu and Yu, Haofei and Zhao, Yining and Liu, Ge and McAuley, Julian and You, Jiaxuan},
  journal = {arXiv preprint arXiv:2605.20616},
  year    = {2026},
  doi     = {10.48550/arXiv.2605.20616}
}

@inproceedings{rag,
  title     = {Retrieval-Augmented Generation for Knowledge-Intensive {NLP} Tasks},
  author    = {Lewis, Patrick and Perez, Ethan and Piktus, Aleksandra and Petroni, Fabio and Karpukhin, Vladimir and Goyal, Naman and K{\"u}ttler, Heinrich and Lewis, Mike and Yih, Wen-tau and Rockt{\"a}schel, Tim and Riedel, Sebastian and Kiela, Douwe},
  booktitle = {Advances in Neural Information Processing Systems (NeurIPS)},
  year      = {2020},
  doi       = {10.48550/arXiv.2005.11401},
  note      = {arXiv:2005.11401}
}

@inproceedings{dpr,
  title     = {Dense Passage Retrieval for Open-Domain Question Answering},
  author    = {Karpukhin, Vladimir and O{\u{g}}uz, Barlas and Min, Sewon and Lewis, Patrick and Wu, Ledell and Edunov, Sergey and Chen, Danqi and Yih, Wen-tau},
  booktitle = {Proceedings of the 2020 Conference on Empirical Methods in Natural Language Processing (EMNLP)},
  pages     = {6769--6781},
  year      = {2020},
  doi       = {10.18653/v1/2020.emnlp-main.550}
}

@article{bm25,
  title   = {The Probabilistic Relevance Framework: {BM25} and Beyond},
  author  = {Robertson, Stephen E. and Zaragoza, Hugo},
  journal = {Foundations and Trends in Information Retrieval},
  volume  = {3},
  number  = {4},
  pages   = {333--389},
  year    = {2009},
  doi     = {10.1561/1500000019}
}

@inproceedings{raptor,
  title     = {{RAPTOR}: Recursive Abstractive Processing for Tree-Organized Retrieval},
  author    = {Sarthi, Parth and Abdullah, Salman and Tuli, Aditi and Khanna, Shubh and Goldie, Anna and Manning, Christopher D.},
  booktitle = {International Conference on Learning Representations (ICLR)},
  year      = {2024},
  doi       = {10.48550/arXiv.2401.18059},
  note      = {arXiv:2401.18059}
}

@article{graphrag,
  title   = {From Local to Global: A Graph {RAG} Approach to Query-Focused Summarization},
  author  = {Edge, Darren and Trinh, Ha and Cheng, Newman and Bradley, Joshua and Chao, Alex and Mody, Apurva and Truitt, Steven and Metropolitansky, Dasha and Ness, Robert Osazuwa and Larson, Jonathan},
  journal = {arXiv preprint arXiv:2404.16130},
  year    = {2024},
  doi     = {10.48550/arXiv.2404.16130}
}

@article{panini,
  title   = {{Panini}: Continual Learning in Token Space via Structured Memory},
  author  = {Rajesh, S. and Holur, P. and Turali, M. Y. and Duan, C. and Roychowdhury, V.},
  journal = {arXiv preprint arXiv:2602.15156},
  year    = {2026},
  doi     = {10.48550/arXiv.2602.15156},
  note    = {Published at ICML 2026}
}

@article{dspy,
  title   = {{DSPy}: Compiling Declarative Language Model Calls into Self-Improving Pipelines},
  author  = {Khattab, Omar and Singhvi, Arnav and Maheshwari, Paridhi and Zhang, Zhiyuan and Santhanam, Keshav and Vardhamanan, Sri and Haq, Saiful and Sharma, Ashutosh and Joshi, Thomas T. and Moazam, Hanna and Miller, Heather and Zaharia, Matei and Potts, Christopher},
  journal = {arXiv preprint arXiv:2310.03714},
  year    = {2023},
  doi     = {10.48550/arXiv.2310.03714}
}

@article{gepa,
  title   = {{GEPA}: Reflective Prompt Evolution Can Outperform Reinforcement Learning},
  author  = {Agrawal, Lakshya A. and Tan, Shangyin and Soylu, Dilara and Ziems, Noah and Khare, Rishi and Opsahl-Ong, Krista and Singhvi, Arnav and Shandilya, Herumb and Ryan, Michael J. and Jiang, Meng and Potts, Christopher and Sen, Koushik and Dimakis, Alexandros G. and Stoica, Ion and Klein, Dan and Zaharia, Matei and Khattab, Omar},
  journal = {arXiv preprint arXiv:2507.19457},
  year    = {2025},
  doi     = {10.48550/arXiv.2507.19457}
}

@article{promptbreeder,
  title   = {Promptbreeder: Self-Referential Self-Improvement via Prompt Evolution},
  author  = {Fernando, Chrisantha and Banarse, Dylan and Michalewski, Henryk and Osindero, Simon and Rockt{\"a}schel, Tim},
  journal = {arXiv preprint arXiv:2309.16797},
  year    = {2023},
  doi     = {10.48550/arXiv.2309.16797}
}

@inproceedings{appworld,
  title     = {{AppWorld}: A Controllable World of Apps and People for Benchmarking Interactive Coding Agents},
  author    = {Trivedi, Harsh and Khot, Tushar and Hartmann, Mareike and Manku, Ruskin and Dong, Vinty and Li, Edward and Gupta, Shashank and Sabharwal, Ashish and Balasubramanian, Niranjan},
  booktitle = {Proceedings of the 62nd Annual Meeting of the Association for Computational Linguistics (ACL)},
  pages     = {16022--16076},
  year      = {2024},
  doi       = {10.18653/v1/2024.acl-long.850},
  note      = {arXiv:2407.18901}
}

@article{browsecompplus,
  title   = {{BrowseComp-Plus}: A More Fair and Transparent Evaluation Benchmark of Deep-Research Agent},
  author  = {Chen, Zijian and Ma, Xueguang and Zhuang, Shengyao and Nie, Ping and Zou, Kai and Liu, Andrew and Green, Joshua and Patel, Kshama and Meng, Ruoxi and Su, Mingyi and Sharifymoghaddam, Sahel and Li, Yanxi and Hong, Haoran and Shi, Xinyu and Liu, Xuye and Thakur, Nandan and Zhang, Crystina and Gao, Luyu and Chen, Wenhu and Lin, Jimmy},
  journal = {arXiv preprint arXiv:2508.06600},
  year    = {2025},
  doi     = {10.48550/arXiv.2508.06600}
}

@article{dabstep,
  title   = {{DABstep}: Data Agent Benchmark for Multi-step Reasoning},
  author  = {Egg, Alex and Iglesias Goyanes, Martin and Kingma, Friso and Mora, Andreu and von Werra, Leandro and Wolf, Thomas},
  journal = {arXiv preprint arXiv:2506.23719},
  year    = {2025},
  doi     = {10.48550/arXiv.2506.23719}
}

@misc{harveylab,
  title        = {Harvey's Legal Agent Benchmark ({LAB})},
  author       = {Grupen, Niko and Pereyra, Gabe and Pereyra, Julio},
  howpublished = {Harvey. \url{https://github.com/harveyai/harvey-labs}},
  year         = {2026}
}

@article{apexagents,
  title   = {{APEX-Agents}},
  author  = {Vidgen, Bertie and Mann, Austin and Fennelly, Abby and Stanly, John Wright and Rothman, Lucas and Burstein, Marco and Benchek, Julien and Ostrofsky, David and Ravichandran, Anirudh and Sur, Debnil and Venugopal, Neel and Hsia, Alannah and Robinson, Isaac and Huang, Calix and Varones, Olivia and Khan, Daniyal and Haines, Michael and Bridges, Austin and Boyle, Jesse and Twist, Koby and Richards, Zach and Mahapatra, Chirag and Foody, Brendan and Nitski, Osvald},
  journal = {arXiv preprint arXiv:2601.14242},
  year    = {2026},
  doi     = {10.48550/arXiv.2601.14242},
  note    = {The AI Productivity Index for agents}
}

@article{officeqa,
  title   = {{OfficeQA Pro}: An Enterprise Benchmark for End-to-End Grounded Reasoning},
  author  = {Opsahl-Ong, Krista and Singhvi, Arnav and Collins, Jasmine and Zhou, Ivan and Wang, Cindy and Baheti, Ashutosh and Oertell, Owen and Portes, Jacob and Havens, Sam and Elsen, Erich and Bendersky, Michael and Zaharia, Matei and Chen, Xing},
  journal = {arXiv preprint arXiv:2603.08655},
  year    = {2026},
  doi     = {10.48550/arXiv.2603.08655},
  note    = {Benchmark suite: \url{https://github.com/databricks/officeqa}}
}
